\documentclass{article}

\usepackage{microtype}
\usepackage{graphicx}
\usepackage{booktabs} 
\usepackage{subcaption,caption}

\usepackage{amsmath}
\usepackage{amssymb}
\usepackage{amsfonts}
\usepackage{bm}
\usepackage{tabularx}
\usepackage{float}
\usepackage{wrapfig}

\usepackage{hyperref}

\usepackage[accepted]{icml2019}

\icmltitlerunning{Differentiable Physics-informed Graph Networks}

\begin{document}
\newcommand{\eat}[1]{}
\newcommand{\ryedit}[1]{\textcolor{blue}{\emph{[RY: #1]}}}
\newcommand{\cxbedit}[1]{\textcolor{cyan}{\emph{[CXB: #1]}}}
\newcommand{\tbedit}[1]{\textcolor{magenta}{\emph{[TB: #1]}}}
\newcommand{\ssedit}[1]{{\color{blue} #1}}
\newcommand{\sscomment}[1]{\ssedit{[SS: #1]}}
\newcommand{\mtedit}[1]{{\color{blue} #1}}
\newcommand{\mtcomment}[1]{\mtedit{[MT: #1]}}
\newcommand{\todo}[1]{\textcolor{red}{\emph{TODO: #1}}}
\newcommand{\red}[1]{\textcolor{red}{#1}}

\newenvironment{example}[1][Example]{\begin{trivlist}
\item[\hskip \labelsep {\bfseries #1}]}{\end{trivlist}}
\newenvironment{remark}[1][Remark]{\begin{trivlist}
\item[\hskip \labelsep {\bfseries #1}]}{\end{trivlist}}

\newtheorem{definition}{Definition}
\newtheorem{problem}{Problem}
\newtheorem{theorem}{Theorem}[section]
\newtheorem{lemma}[theorem]{Lemma}
\newtheorem{note}[theorem]{Note}
\newtheorem{corollary}[theorem]{Corollary}
\newtheorem{prop}[theorem]{Proposition}

\newcommand{\abs}[1]{ {\left| #1 \right|}}
\newcommand{\pq}[1]{\left( #1 \right)}
\newcommand{\cpd}[1]{{\left\llbracket {#1} \right\rrbracket}}

\newcommand{\mb}[1]{{\mathbf{#1}}}
\newcommand{\V}[1]{{\bm{#1}}} 
\newcommand{\M}[1]{{\bm{#1}}} 
\newcommand{\T}[1]{{\mb{\mathsf{#1}}}} 
\newcommand{\C}[1]{\mathcal{#1}} 

\newcommand{\adj}{\M{A}}
\newcommand{\gweights}{{\M{W}}}
\newcommand{\ex}{\mathbb{E}}
\newcommand{\edges}{\C{E}}
\newcommand{\graph}{\C{G}}
\newcommand{\vertices}{\C{V}}
\newcommand{\real}{\mathbb{R}}
\newcommand{\skt}{{\mathcal{S}}}
\newcommand{\lapl}{\M{L}}
\newcommand{\loss} {\mathcal{L}}
\newcommand{\ppr}{{\C{P}}}  

\newcommand{\hprod}{\odot}
\newcommand{\kprod}{\otimes}
\newcommand{\krprod}{\odot}
\newcommand{\oprod}{\circ }
\newcommand{\gconv}{{\,\star_{\mathcal{G}}\,}}
\newcommand{\spectral}[1]{\widehat{#1}}

\newcommand{\SM}[1]{\mathbf{\hat{#1}}}
\newcommand{\transpose}{\intercal}
\newcommand{\bestval}[1]{{\textbf{#1}}}

\newcommand{\SV}[2]{\sigma_{#1}(#2)}
\newcommand{\EV}[2]{\lambda_{#1}(#2)}
\newcommand{\numberthis}{\addtocounter{equation}{1}\tag{\theequation}}

\newcommand{\ours}{{DPGN}} 
\newcommand{\dummyfigure}[2]{\fbox{\begin{minipage}{#1}\hfill\vspace{#2}\end{minipage}}}

\twocolumn[
\icmltitle{Differentiable Physics-informed Graph Networks}




\begin{icmlauthorlist}
\icmlauthor{Sungyong Seo}{to}
\icmlauthor{Yan Liu}{to}
\end{icmlauthorlist}

\icmlaffiliation{to}{Department of Computer Science, University of Southern California, Los Angeles, USA}

\icmlcorrespondingauthor{Sungyong Seo}{sungyons@usc.edu}


\vskip 0.3in
]



\printAffiliationsAndNotice{}  

\begin{abstract}
While physics conveys knowledge of nature built from an interplay between observations and theory, it has been considered less importantly in deep neural networks.
Especially, there are few works leveraging physics behaviors when the knowledge is given less explicitly. 
In this work, we propose a novel architecture called Differentiable Physics-informed Graph Networks ({\ours}) to incorporate implicit physics knowledge which is given from domain experts by informing it in latent space.
Using the concept of {\ours}, we demonstrate that climate prediction tasks are significantly improved. 
Besides the experiment results, we validate the effectiveness of the proposed module and provide further applications of {\ours}, such as inductive learning and multistep predictions.
\end{abstract}

\section{Introduction}\label{sec:intro}

Modeling natural phenomena in the real-world, such as climate, traffic, molecule, and so on, is extremely challenging but important. 
Deep learning has achieved significant successes in prediction performance by learning latent representations from \textit{data-rich} applications such as speech recognition~\cite{hinton2012deep}, text understanding~\cite{wu2016google}, and image recognition~\cite{krizhevsky2012imagenet}.  
While the accuracy and efficiency of data-driven deep learning models can be improved with ad-hoc architectural changes for specific tasks, we are confronted with many challenging learning  scenarios in modeling natural phenomenon, where a limited number of labeled examples are available or there is much noise in the data. Furthermore, there could be constant changes in data distributions (e.g. dynamic systems). Therefore, there is a pressing need to develop new generation ``deeper'' and robust learning models that can  address these challenging learning scenarios.

Physics is one of the fundamental pillars describing how the real-world behaves. 
It is imperative that physics-informed learning models are powerful solutions to modeling natural phenomena.  
Incorporating domain knowledge has several benefits: first, it helps an optimized solution to be more stable and to prevent overfitting; second, it provides theoretical guidance with which an effective model is supposed to follow and thus, helps training with less data; lastly, since a model is driven by the desired knowledge, it would be more robust to unseen data, and thus it is easier to be extended to applications with changing distributions.

\begin{figure}[t]
    \centering
    \includegraphics[width=0.75\linewidth]{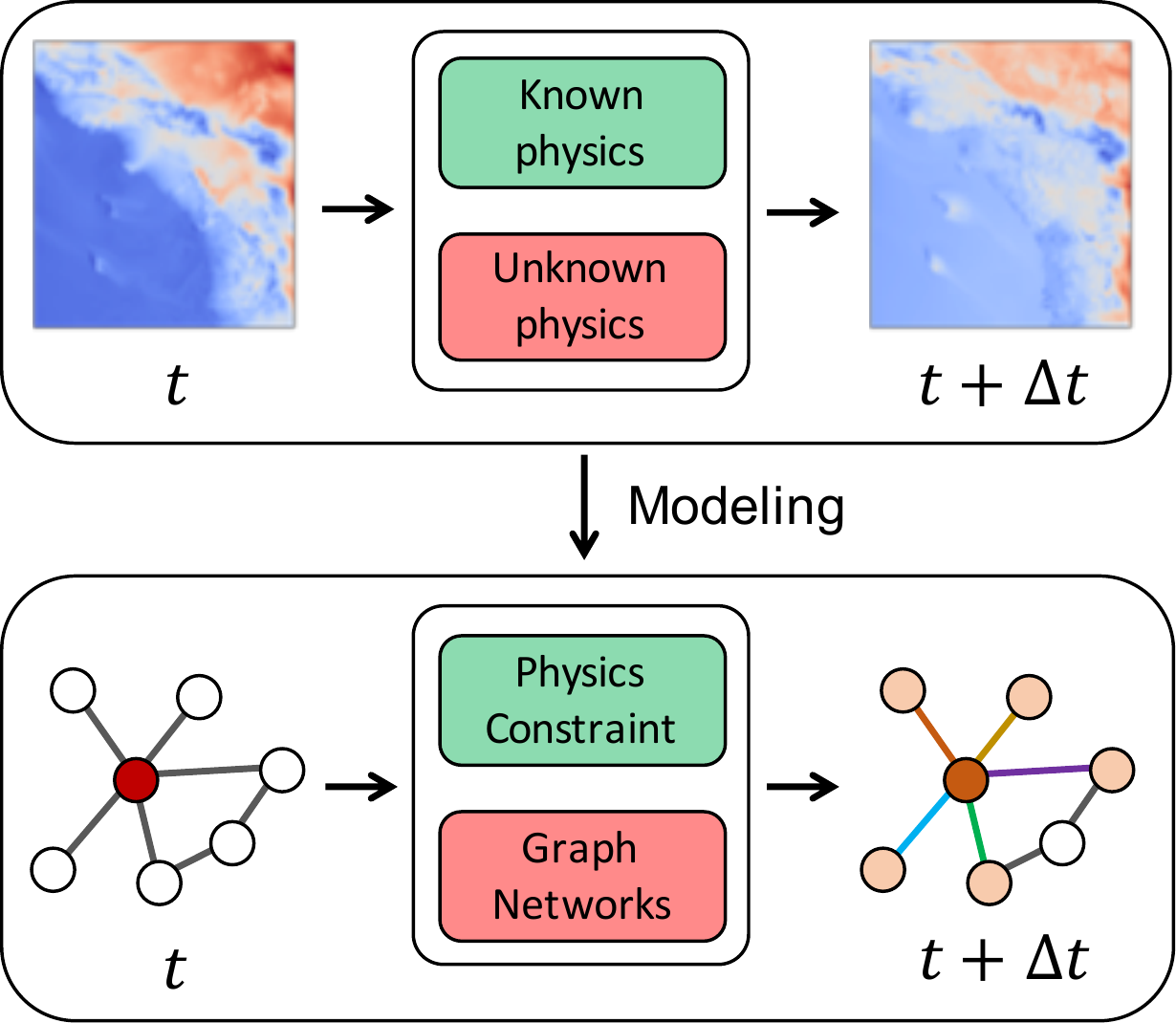}
    \caption{Concept of the proposed {\ours}. The behaviors of sequential observations (Temperature) are governed by physics rules. Some of the physics rules are known and we inject them into a model explicitly. The remained unknown patterns will be extracted from data.}
    \label{fig:concept}
    \vskip -0.2in
\end{figure}

In the meanwhile, there exist a series of challenges when we incorporate physics principles into machine learning models. 
First, a model needs to be able to properly handle the spatial and temporal constraints. 
Many physics equations demonstrate how a set of physical quantities behaves over time and space.
For example, the wave equation describes how a signal is propagated through a medium over time.
Second, the model should capture relations between objects, such as image patches~\cite{santoro2017simple} or rigid bodies~\cite{battaglia2016interaction,chang2016compositional}.
Third, the learning modules should be common for all objects because physical phenomena apply to all objects.
Finally, the model should be flexible to extract unknown patterns instead of being strictly constrained to physics knowledge.
Since it is not always possible to describe all rules governing real-world data, data-driven learning is required to fill the gap between the known physics and real observations.

In this paper, we address the problem of modeling dynamical systems based on graph-based neural networks by incorporating useful knowledge described as differentiable physics equations.
We propose a generic architecture, differentiable physics-informed graph networks (\ours), which can leverage explicitly required physics and learn implicit patterns from data as illustrated in Figure~\ref{fig:concept}.
The proposed model properly handles spatially located objects and their relations as vertices and edges in a graph. 
Moreover, temporal dependencies are learned by recurrent computations.
As \citet{battaglia2018relational} suggest, the inductive bias of a graph-based model is its \emph{invariance [to] node/edge permutations}, and thus, all trainable functions for the same input types are shared.

Our contributions of this work are summarized as follows:
\begin{itemize}
    \item We develop a novel physics-informed learning architecture, {\ours}, which incorporates differentiable physics equations with a graph network framework.
    \item We develop a variant of {\ours} so that it can infer latent patterns that cannot be captured by existing physics knowledge but is important to model observations.
    \item We investigate the effectiveness of {\ours} for climate modeling in terms of prediction and inductive learning. Moreover, we illustrate how physics knowledge can help improve prediction performance.
\end{itemize}
    
The rest of this paper is organized as follows. 
Section~\ref{sec:related} describes related work, 
Section~\ref{sec:background} provides the basic mathematics of differential operators in a graph. The proposed idea is presented in Section~\ref{sec:pign} and Section~\ref{sec:exp} reports numerical experiments and analysis.

\section{Related Work}\label{sec:related}


\paragraph{Incorporating physics} Many attempts have been made on incorporating physical knowledge into data-driven models.
\citet{cressie2015statistics} covered a number of statistical models handling physical equations. A hierarchical Bayesian framework is mainly used for modeling real-world data in this work. 
\citet{raissi2017physicsI,raissi2017physicsII} introduced a concept of physics-informed neural networks, which utilize physics equations explicitly to train neural networks. 
By optimizing the model at initial/boundary data and sampled collocation points, the data-driven solutions of nonlinear PDEs can be found. Based on this fundamental idea, a number of works for simulating and discovering PDEs have been published~\cite{raissi2018hidden,raissi2018deep}.

Although these works leveraged physical knowledge, they are limited because they require all physics behind given data to be explicitly known.
\citet{de2018deep} considered a similar problem as ours. 
They proposed how transport physics (advection and diffusion) could be incorporated for forecasting sea surface temperature (SST). 
Namely, they proposed how the motion flow that is helpful for the temperature flow prediction could be extracted in an unsupervised manner from a sequence of SST images.
This work is a major milestone since it captures not only the dominant transport physics but also unknown patterns inferred through the neural networks.
Despite of its novel architecture, the model is specifically designed for transport physics and it is not straightforward to extend the model to other physics equations. 
Furthermore, it is restricted in a regular grid since it used conventional convolutional neural networs (CNNs) for images.

\paragraph{Learning physical dynamics} A class of models~\cite{grzeszczuk1998neuroanimator,battaglia2016interaction,chang2016compositional,watters2017visual,pmlr-v80-sanchez-gonzalez18a,kipf2018neural} have been proposed based on the assumption that neural networks can learn complex physical interactions and simulate unseen dynamics based on a current state. 
The models along this direction are based on common \textit{relational inductive biases}~\cite{santoro2017simple,battaglia2018relational}, i.e., functions connecting entities and relations are shared and can be learned from a given sequence of simulated dynamics.
\citet{chang2016compositional,battaglia2016interaction,pmlr-v80-sanchez-gonzalez18a} commonly assumed that the objects' behaviors were governed by classical kinetic physics equations.
Then, object- and relation-centric functions were proposed to learn the transition from the current state to the next state without explicitly injecting the equations into the model.
Unlike this line of works that implicitly extracts latent patterns from data only, our proposed model can incorporate known physics and at the same time extract latent patterns from data which cannot be captured by existing knowledge.


\section{Background}\label{sec:background}
Since the concept of graph-based neural networks was first proposed in~\citet{scarselli2009graph}, a number of variants have been studied.
These works commonly focus on how to propagate \emph{latent representations} of nodes and edges (or even a whole graph) to obtain better representations of them.
One branch of these studies is a spectral approach to aggregate locally connected attributes~\cite{bruna2013spectral,defferrard2016convolutional,kipf2016semi} and another branch is a spatial approach (or message passing method)~\cite{gilmer2017neural,monti2017geometric,hamilton2017inductive}.
Although these works have different views, it is possible to describe each work as a special case of Graph Networks (GN)~\cite{battaglia2018relational} because all of them share the common inductive bias, \emph{node/edge permutation invariance}.
In this section, we introduce how operators of vector calculus in Euclidean domain are analogously defined on the discrete graph domain and briefly show that the graph networks module is able to efficiently express the differential operators.


\subsection{Calculus on Graphs}
Vector calculus is concerned with differentiation of vector/scalar fields primarily in 3-dimensional Euclidean space $\mathbb{R}^3$. 
It has an essential role in partial differential equations (PDEs) in physics and engineering.

\begin{figure}[t]
    \centering
    \begin{subfigure}{0.4\linewidth}
        \centering
        \includegraphics[width=0.7\linewidth]{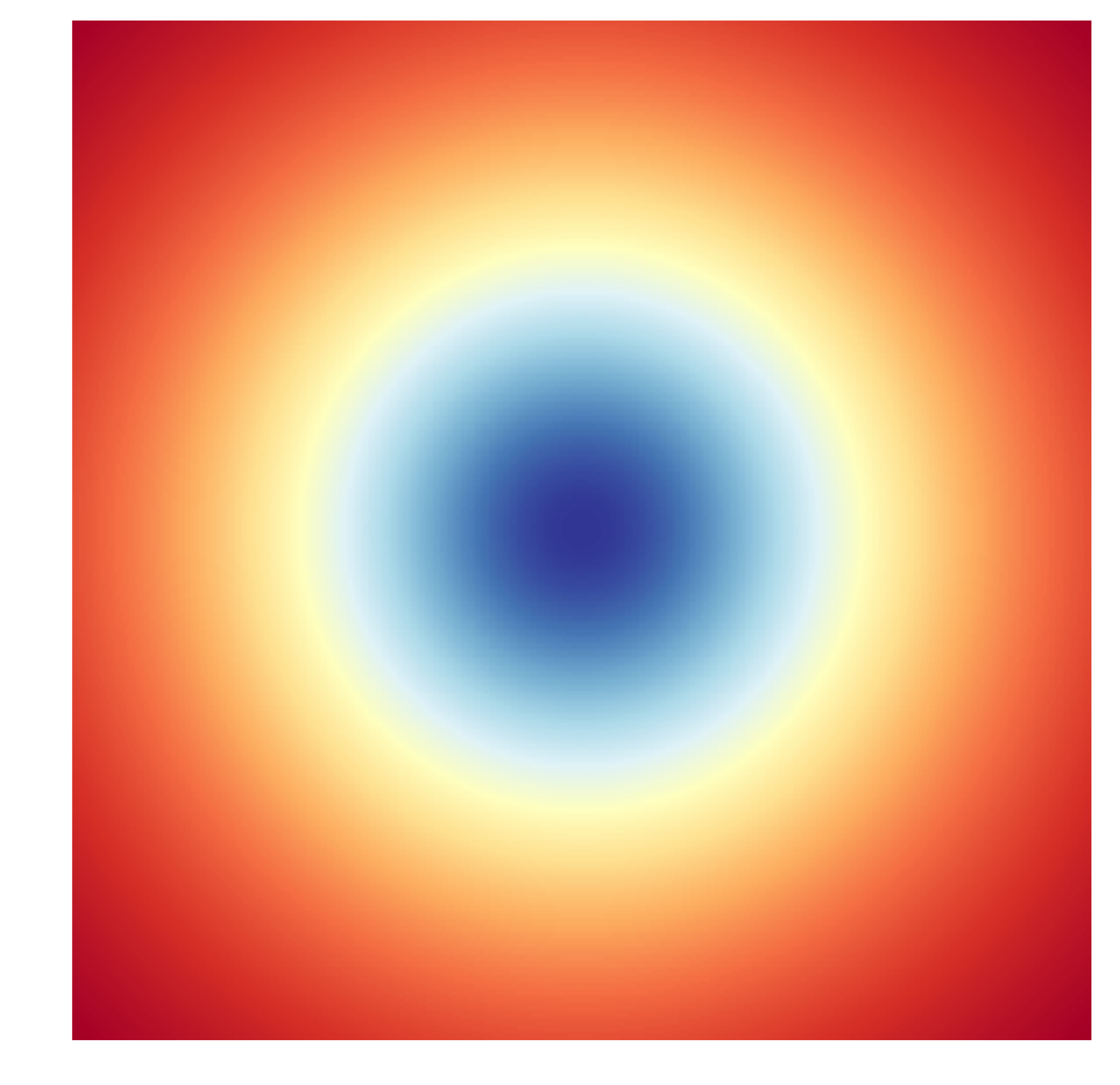}
        \caption{Scalar field}
    \end{subfigure}
    \begin{subfigure}{0.4\linewidth}  
        \centering 
        \includegraphics[width=0.7\linewidth]{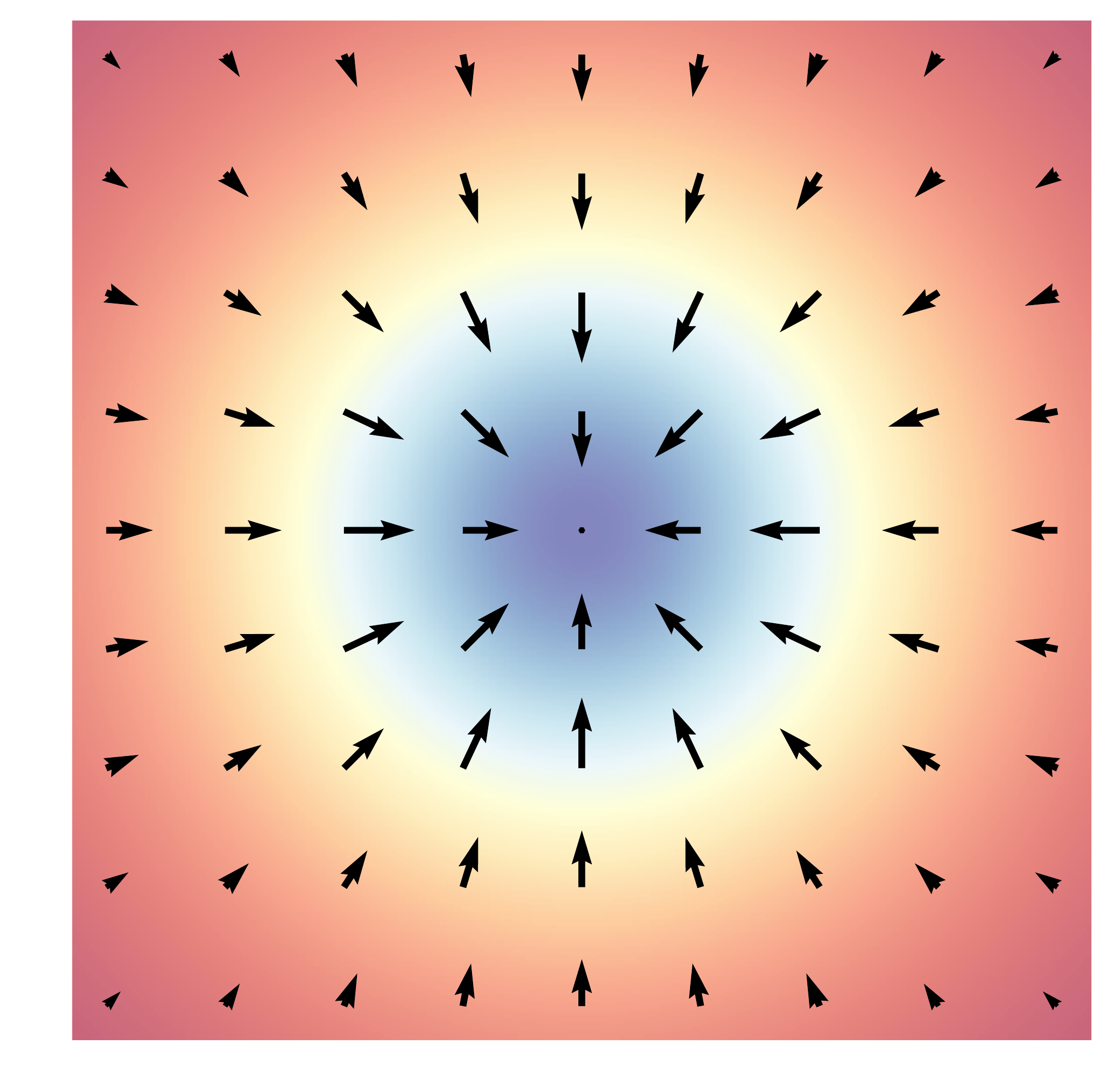}
        \caption{Vector field}
    \end{subfigure}
    \vskip\baselineskip
    \begin{subfigure}{0.4\linewidth}   
        \centering 
        \includegraphics[width=0.7\linewidth]{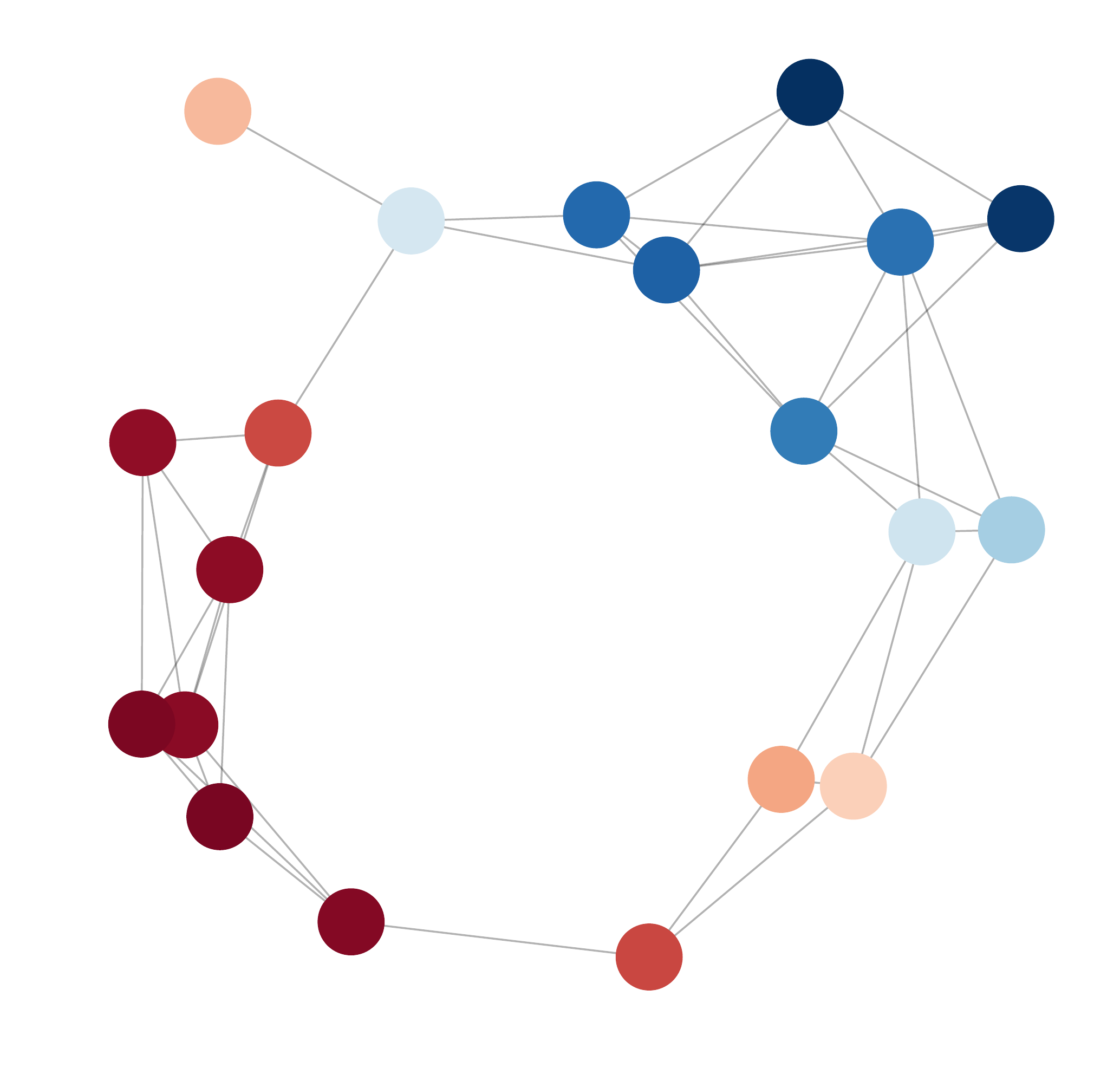}
        \caption{Vertex function}
    \end{subfigure}
    \begin{subfigure}{0.4\linewidth}   
        \centering 
        \includegraphics[width=0.7\linewidth]{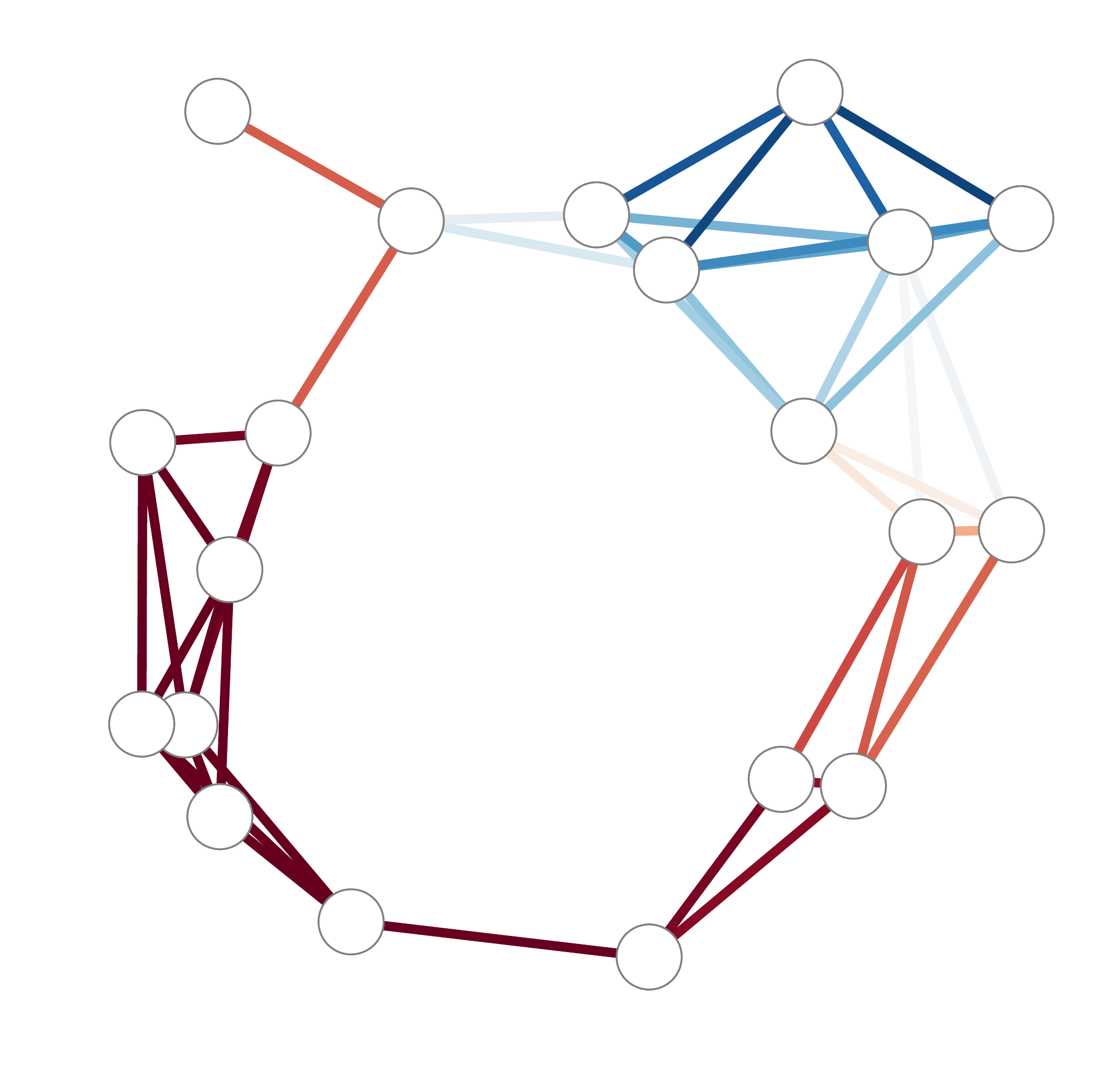}
        \caption{Edge function}
    \end{subfigure}
    \caption{Scalar/vector fields on Euclidean space and vertex/edge functions on a graph.} 
    \label{fig:fields}
    \vskip -0.2in
\end{figure}

\paragraph{Preliminary} Given a graph $\C{G}=(\C{V},\C{E})$ where $\C{V}$ and $\C{E}$ are a set of vertices $\C{V}=\{1,\dots,n\}$ and edges $\C{E}\subseteq\bigl(\begin{smallmatrix}\C{V} \\ 2\end{smallmatrix}\bigr)$, respectively, two types of real functions can be defined on the vertices, $f:\C{V}\rightarrow\mathbb{R}$, and edges, $F:\C{E}\rightarrow\mathbb{R}$, of the graph.
It is also possible to define multiple functions on the vertices or edges as multiple feature maps of a pixel in CNNs.
Since $f$ and $F$ can be viewed as scalar fields and tangent vector fields in differential geometry (Figure~\ref{fig:fields}), the corresponding discrete operators on graphs can be defined as follow~\cite{lim2015hodge}.

\paragraph{Gradient on graphs} The \emph{gradient} on a graph is the linear operator defined by
\begin{align*}
& \nabla:L^2(\C{V})\rightarrow L^2(\C{E})\\
& (\nabla f)_{ij}=(f_j-f_i)\quad\text{if }\{i,j\}\in\C{E}\text{ and 0 otherwise.}
\end{align*}
where $L^2(\C{V})$ and $L^2(\C{E})$ denote Hilbert spaces of vertex functions and edge functions, respectively, thus $f\in L^2(\C{V})$ and $F\in L^2(\C{E})$. As the gradient in Euclidean space measures the rate and direction of change in a scalar field, the gradient on a graph computes \emph{differences} of the values between two adjacent vertices and the differences are defined along the directions of the corresponding edges.

\paragraph{Divergence on graphs} The \emph{divergence} in Euclidean space maps vector fields to scalar fields. Similarly, the divergence on a graph is the linear operator defined by
\begin{align*}
& \text{div}:L^2(\mathcal{E})\rightarrow L^2(\mathcal{V})\\
& (\text{div }F)_i=\sum_{j:(i,j)\in\mathcal{E}}w_{ij}F_{ij}\quad\forall{i}\in\C{V}
\end{align*}
where $w_{ij}$ is a weight on the edge $(i,j)$.
It denotes a weighted sum of incident edge functions to a vertex $i$, which is interpreted as the netflow at a vertex $i$.

\paragraph{Laplacian on graphs} \emph{Laplacian} ($\Delta=\nabla^2)$ in Euclidean space measures the difference between the values of the scalar field with its average on infinitesimal balls. Similarly, the graph Laplacian is defined as
\begin{align*}
    & \Delta:L^2(\C{V})\rightarrow L^2(\C{V})\\
    & (\Delta f)_i = \sum_{j:(i,j)\in\mathcal{E}}w_{ij}(f_i-f_j)\quad\forall{i}\in\C{V}
\end{align*}
The graph Laplacian can be represented as a matrix form, $L=D-W$ where $D=\text{diag}(\sum_{j:j\neq i}w_{ij})$ is a degree matrix and $W$ denotes a weighted adjacency matrix. Note that $L=\Delta=-\text{div}\nabla$ and the minus sign is required to make $L$ positive semi-definite.

\paragraph{Curl on graphs} The \emph{curl} on a graph is a more complicated concept defined on 3-cliques, the set of triangles $\C{T}\subseteq\bigl(\begin{smallmatrix}\C{V} \\ 3\end{smallmatrix}\bigr)$ where $\{i,j,k\}\in\C{T}$ if and only if $\{i,j\},\{i,k\},\{j,k\}\in\C{E}$.
\begin{align*}
    &\text{curl}:L^2(\C{E})\rightarrow L^2(\C{T})\\
    &(\text{curl }F)(i,j,k) = F(i,j) + F(j,k) + F(k,i)\\
    &\text{curl*}:L^2(\C{T})\rightarrow L^2(\C{E})\\
    &(\text{curl* }\mathtt{F})(i,j) = \sum_{k}\frac{w_{ijk}}{w_{ij}}\mathtt{F}(i,j,k)\\
    & \qquad\qquad\qquad\text{if }\{i,j,k\}\in\C{T}\text{ and 0 otherwise.}
\end{align*}
where $\mathtt{F}(\cdot)\in L^2(\C{T})$ is a 3-clique function and $w_{ijk}$ is a weight defined on the clique $\{i,j,k\}$. curl* is an expression for the adjoint of the curl operator. 
Note that some vector calculus properties (e.g., solenoidal or irrotational) can be seamlessly verified (div curl* $F=0$ or curl $\nabla f=0$).

Based on the core differential operators on a graph, we can re-write differentiable physics equations (e.g., Diffusion equation or Wave equation) on a graph.


\subsection{Graph Networks}
\citet{battaglia2018relational} proposed a graph networks framework, which generalizes relations among vertices, edges, and a whole graph.
Graph networks describe how edge, node, and global attributes are updated by propagating information among themselves.

Given a set of nodes ($\bm{v}$), edges ($\bm{e}$), and global ($\bm{u}$) attributes, the steps of computation in a graph networks block are as follow:
\begin{enumerate}
    \item $\bm{e}'_{ij}\leftarrow\phi^e(\bm{e}_{ij},\bm{v}_i,\bm{v}_j,\bm{u})$ for all $\{i,j\}\in\C{E}$ pairs.
    \item $\bm{v}'_i\leftarrow\phi^v(\bm{v}_i,\bar{\bm{e}}'_i,\bm{u})$ for all $i\in\C{V}$.\\
    $\bar{\bm{e}}'_i$ is an aggregated edge attribute related to the node $i$.
    \item $\bm{u}'\leftarrow\phi^u(\bm{u},\bar{\bm{e}}',\bar{\bm{v}}')$\\
    $\bar{\bm{e}}'$ and $\bar{\bm{v}}'$ are aggregated attributes of all edges and all nodes in a graph, respectively.
\end{enumerate}
where $\phi^e, \phi^v, \phi^u$ are edge, node, and global update functions, respectively, and they can be implemented by learnable feed forward neural networks.
Note that the computation order is flexible. The aggregators can be chosen freely once it is invariant to permutations of their inputs. Furthermore, a set of input for each mapping function can be also customized.

As $\phi^e$ is a mapping function from vertices to edges, it can be replaced by the gradient operator to describe the known relation explicitly (e.g., Electric potential $\rightarrow$ Electric field).
Similarly, $\phi^v$ can learn divergence-like mapping functions.
For curl-involved functions, it is required to add another updating function, $\phi^c$, which is mapping from nodes/edges/global attributes to a 3-clique attribute and vice versa.
In other words, the graph networks have highly flexible modules which are able to imitate the differential operators in a graph explicitly or implicitly.


\section{Differentiable Physics-informed Graph Networks}\label{sec:pign}
As deep learning models are successful to model complex behaviors or extract abstract features in real-world data, it is natural to focus on how the data-driven modeling can solve practical problems in physics or engineering fields.
In this section, we provide how useful information described in physics processes can be incorporated with the graph networks framework.

\subsection{Static Physics}\label{sec:static}
Many fields in physics dealing with static properties, such as Electrostatic, Magnetostatic, or Hydrostatic, describe a number of physics phenomena at rest.
Among the various phenomena, it is easy to express differentiable physics rules in discrete forms on a graph with the operators in Section~\ref{sec:background}.
For instances, the Poisson equation ($\nabla^2\phi=-\frac{\rho}{\epsilon_0}$) in Electrostatics is realized as a simple matrix multiplication of graph Laplacian with a vertex function. 
In other words, the static physics can be modeled by replacing the updating functions in graph networks with corresponding operators.
Table~\ref{tab:static} provides some differential formulas in Electrostatic and how the updating functions are defined in GN.
\begin{table}[h]
\centering
\caption{Examples of static physics and updating functions in GN}\label{tab:static}
\vskip 0.1in
\begin{small}
  \begin{tabular}{lcc}
    Mapping         & Updating function & Physics example   \\ \cline{1-3}
\begin{tabular}[c]{@{}l@{}}node\\$\rightarrow$ edge\end{tabular} & \parbox[c][1.2cm]{1cm}{\begin{align*}
                \bm{e}_{ij} &=\phi^e(\bm{v}_i,\bm{v}_j)\\
                &=(\nabla \bm{v})_{ij}
            \end{align*}}
            &   \begin{tabular}[c]{@{}l@{}}$\nabla\phi=-E$\\(Electric field)\end{tabular}        \\
\begin{tabular}[c]{@{}l@{}}edge\\$\rightarrow$ node\end{tabular} &  \parbox[c][1.2cm]{1cm}{\begin{align*}
                \bm{v}_{i} &=\phi^v(\bm{e}_{ij})\\
                &=(\text{div } \bm{e})_i\end{align*}}
            &  \begin{tabular}[c]{@{}l@{}}$\nabla\cdot E=\rho/\epsilon_0$\\(Maxwell's eqn.)\end{tabular}         \\
\begin{tabular}[c]{@{}l@{}}node\\$\rightarrow$ node\end{tabular} &  \parbox[c][1.2cm]{1cm}{\begin{align*}
                \bm{v}_{i} &=\phi^v(\bm{v}_{i}, \{\bm{v}_{j:(i,j)\in\C{E}}\})\\
                &=(\Delta  \bm{v})_i\end{align*}}
            &  \begin{tabular}[c]{@{}l@{}}$\Delta \phi=0$\\(Laplace's eqn.)\end{tabular}         \\ 
\begin{tabular}[c]{@{}l@{}}edge\\$\rightarrow$ 3-clique\end{tabular} &  \parbox[c][1.2cm]{1cm}{\begin{align*}
                \bm{c}_{ijk} &=\phi^c(\bm{e}_{ij},\bm{e}_{jk},\bm{e}_{ki})\\
                &=(\text{curl } \bm{e})_{ijk}\end{align*}}
            &  \begin{tabular}[c]{@{}l@{}}$\nabla\times E=0$\\(Irrotational E-field)\end{tabular}         \\ \cline{1-3}
  \end{tabular}
  \end{small}
\end{table}

\subsection{Dynamic Physics}\label{sec:dynamic}
More practical problems have been written in the dynamic forms, which describe how a given physical quantity is changed in a given region over time.
GN can be regarded as a module that updates a graph state including the attributes of node, edge, 3-clique, and a whole graph. 
\begin{align}
    \C{G}' = \text{GN}(\C{G})\label{eq:update}
\end{align}
where $\C{G}'$ is the updated next graph state.
Many dynamic physics formulas are written as a relation between time derivatives and spatial derivatives:
\begin{align}
    \sum_{i=1}^M\frac{\partial^i u}{\partial t^i}=\sum_{i=1}^Na_i(u,x)\frac{\partial^i u}{\partial x^i}\label{eq:dynamic}
\end{align}
where $u$ is a physical quantity following above PDE and $x$ is the direction where $u$ is defined on. $a_i$ are corresponding coefficients. $M$ and $N$ denote the highest order of time derivatives and spatial derivatives, respectively. 

Based on the state updating view in Equation~\ref{eq:update}, any types of PDEs written in Equation~\ref{eq:dynamic} can be represented as forms of finite differences.
Table~\ref{tab:dynamic} provides the examples of the dynamic physics. $\dot{u}$ and $\ddot{u}$ are the first and second order time derivatives, respectively.

\begin{figure*}[t]
    \centering
    \includegraphics[width=0.7\linewidth]{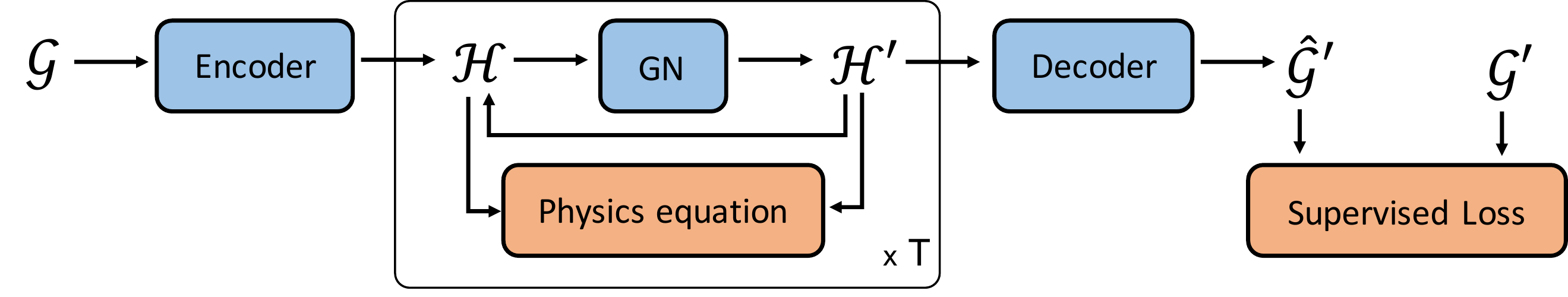}
    \caption{Recurrent architecture to incorporate physics equation on GN. The blue blocks have learnable parameters and the orange blocks are objective functions. The middle core block can be repeated as many as the required time steps ($T$).}
    \label{fig:architecture}
\end{figure*}

\begin{table}[h]
\centering
\caption{Examples of dynamic physics in GN}\label{tab:dynamic}
\vskip 0.1in
\begin{small}
  \begin{tabular}{ll}
    Updating function & Physics example   \\ \cline{1-2}
 \parbox[c][1.2cm]{1cm}{\begin{align*}
                \bm{v}'_i &=\bm{v}_i + \alpha\phi^v(\bm{v}_{i}, \{\bm{v}_{j:(i,j)\in\C{E}}\})\\
                &=\bm{v}_i + \alpha(\Delta\bm{v})_i
            \end{align*}}
            &   \begin{tabular}[c]{@{}l@{}}$\dot{u}=\alpha\nabla^2u$\\(Diffusion eqn.)\end{tabular}        \\
\parbox[c][1.2cm]{1cm}{\begin{align*}
                \bm{v}''_i &=2\bm{v}'_i-\bm{v}_i + c^2\phi^v(\bm{v}'_{i}, \{\bm{v}'_{j:(i,j)\in\C{E}}\})\\
                &=2\bm{v}'_i - \bm{v}_i + c^2(\Delta\bm{v}')_i
            \end{align*}}
            &   \begin{tabular}[c]{@{}l@{}}$\ddot{u}=c^2\nabla^2u$\\(Wave eqn.)\end{tabular}        \\ \cline{1-2}
  \end{tabular}
  \end{small}
\end{table}

\subsection{Physics in Latent Space}
In Section~\ref{sec:static} and~\ref{sec:dynamic}, we provide how the differential operators are implemented in a GN module. 
However, it is hardly practical for modeling complicated real-world problems with the operators solely because it is only possible when all physics equations governing the observed phenomena are explicitly known.
For example, although we understand that there are a number of physics equations involved in climate observations, it is almost infeasible to include all required equations for modeling the observations.
Thus, it is necessary to utilize the learnable parameters in GN to extract latent representations in the data. 
Then, some known physics knowledge will be incorporated with the latent representations.

\paragraph{Forward/Recurrent computation} Figure~\ref{fig:architecture} provides how the desired physics knowledge is integrated with the learnable GN. Given a graph $\C{G}=\{\bm{v},\bm{e},\bm{c},\bm{u}\}$, it is fed into an encoder which transforms a set of attributes of nodes ($\bm{v}$), edges ($\bm{e}$), 3-cliques ($\bm{c}$), and a whole graph ($\bm{u}$) into latent spaces.
\begin{align}
    \tilde{\bm{v}},\tilde{\bm{e}},\tilde{\bm{c}},\tilde{\bm{u}} = \text{Encoder}(\bm{v},\bm{e},\bm{c},\bm{u})\label{eq:enc}
\end{align}
After the encoder, the encoded graph $\C{H}=\{\tilde{\bm{v}},\tilde{\bm{e}},\tilde{\bm{c}},\tilde{\bm{u}}\}$ is repeatedly updated within the core block as many as the required time steps $T$.
For each step, $\C{H}$ is updated to $\C{H}'$ which denotes the next state of the encoded graph.
\begin{align}
    \C{H}' = \text{GN}(\C{H})\label{eq:gn}
\end{align}
Finally, the sequentially updated attributes are re-transformed to the original spaces by a decoder.
\begin{align}
    \bm{v}',\bm{e}',\bm{c}',\bm{u}' = \text{Decoder}(\tilde{\bm{v}}',\tilde{\bm{e}}',\tilde{\bm{c}}',\tilde{\bm{u}}')\label{eq:dec}
\end{align}

\paragraph{Objective functions}
There are two types of objective function in this architecture.
First, we define physics-informed constraints (Equation~\ref{eq:phy1}) between the previous and updated states based on the known knowledge. 
\begin{align}
    &\C{L}_{phy}^i = f_{phy}(\C{H}_i,\C{H}'_{i+1},\cdots,\C{H}'_{i+M-1})\label{eq:phy1}\\
    &\C{L}_{phy} = \sum_{i}\C{L}_{phy}\label{eq:phy2}
^i\end{align}
where $\C{L}_{phy}^i$ is the physics-informed quantity from the input at time step $i$ to the predicted $M-1$ steps.
For example, if we are aware that the observations should have a diffusive property, the diffusion equation can be used as the physics-informed constraint.
\begin{align*}
    f_{phy}(\C{H},\C{H}') = \Vert \bm{v}'-\bm{v} - \alpha\nabla^2\bm{v}\Vert^2
\end{align*}
Secondly, the supervised loss function between the predicted graph, $\hat{\C{G}}'$, and the target graph, $\C{G}'$. This loss function is constructed based on the task, such as the cross-entropy or the mean squared error (MSE).

Finally, the total objective function is a sum of the two constraints:
\begin{align}
    \C{L} = \C{L}_{sup} + \lambda\C{L}_{phy}\label{eq:loss}
\end{align}
where $\lambda$ controls the importance of the physics term.

\section{Experiment}\label{sec:exp}


In this section, we evaluate {\ours} on both a synthetic dataset and a real-world climate dataset over the Southern California region.


\subsection{Synthetic Data}\label{sec:synthetic}
First, we explore whether physics constraints in {\ours} is powerful to infer physics behaviors/patterns.
Here we generate dynamical sequences based on the analytical solutions of two physics equations, a) the diffusion equation and b) the wave equation (See Table~\ref{tab:dynamic}).
For the diffusion equation, we randomly pick a node to assign initially localized heat source and set other nodes with zero values.
For the wave equation, we put a pulse signal at the end of a given path.
Then, we train {\ours} by minimizing the physics constraint term (Equation~\ref{eq:phy2}) only and any single target value is not used to optimize the supervised loss.

\begin{figure}[t]
    \centering
    \begin{subfigure}{0.23\linewidth}
        \centering
        \includegraphics[width=0.8\linewidth]{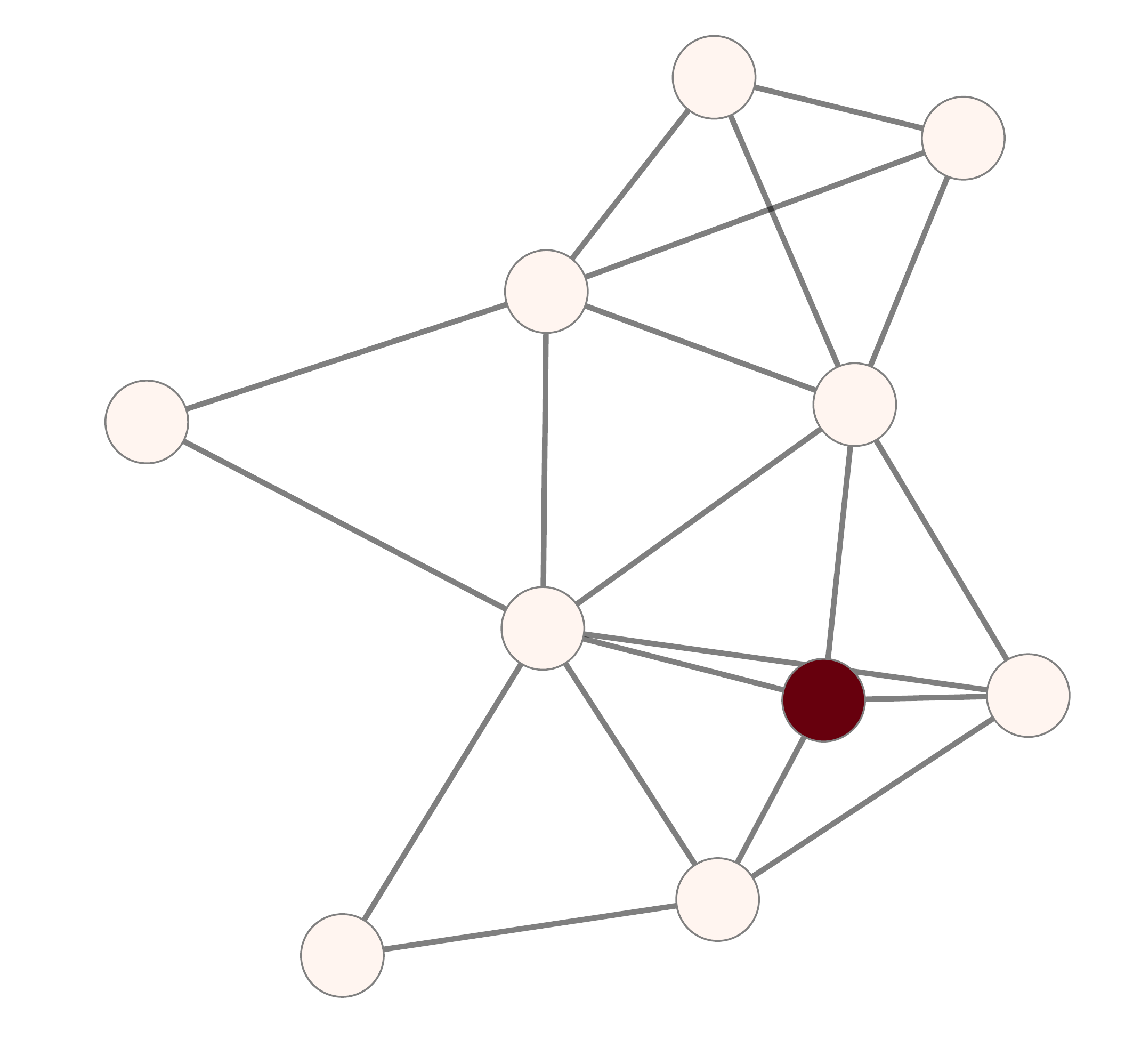}
        \caption*{$t=1$}
    \end{subfigure}
    \begin{subfigure}{0.23\linewidth}  
        \centering 
        \includegraphics[width=0.8\linewidth]{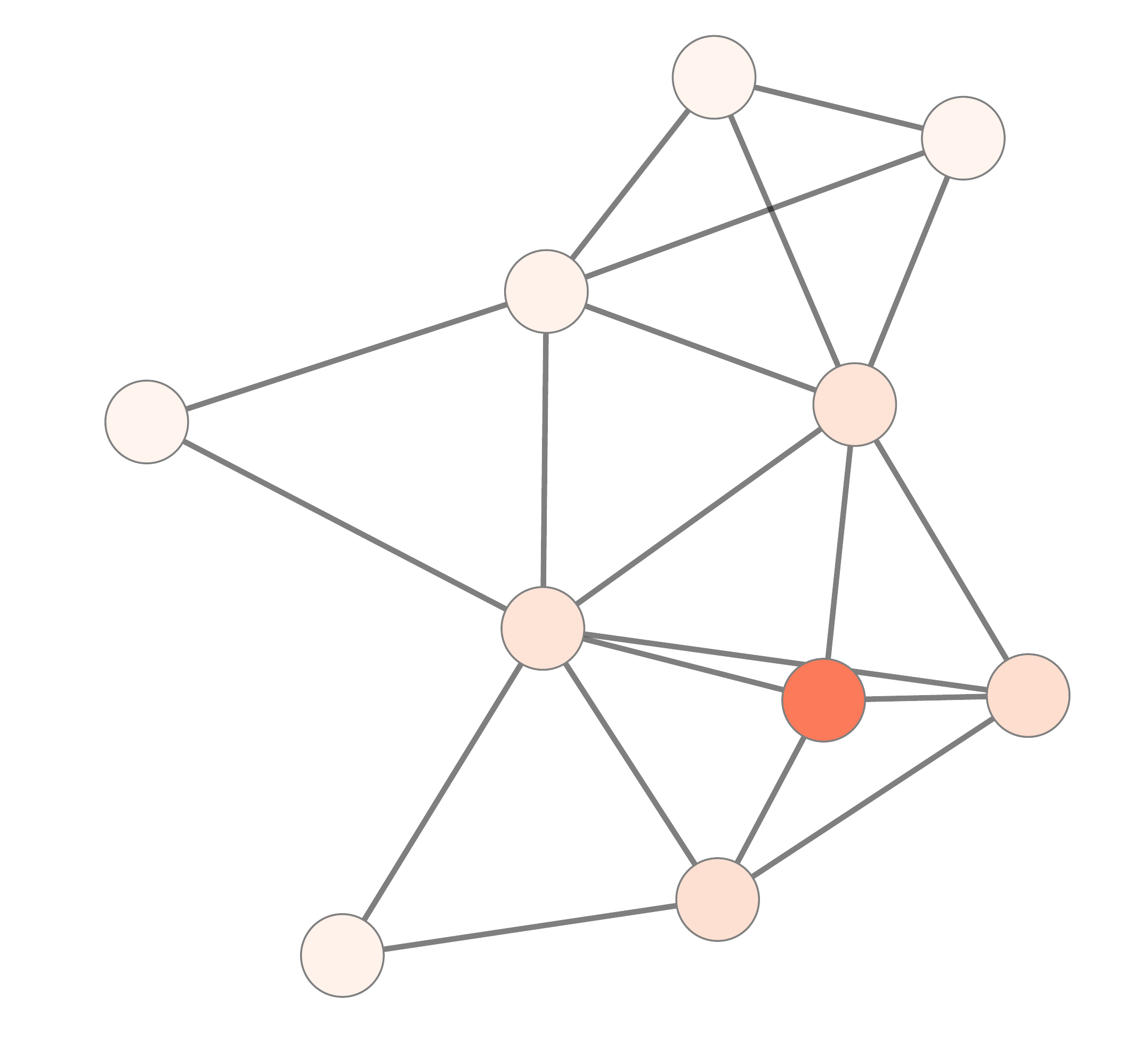}
        \caption*{$t=2$}
    \end{subfigure}
    \begin{subfigure}{0.23\linewidth}  
        \centering 
        \includegraphics[width=0.8\linewidth]{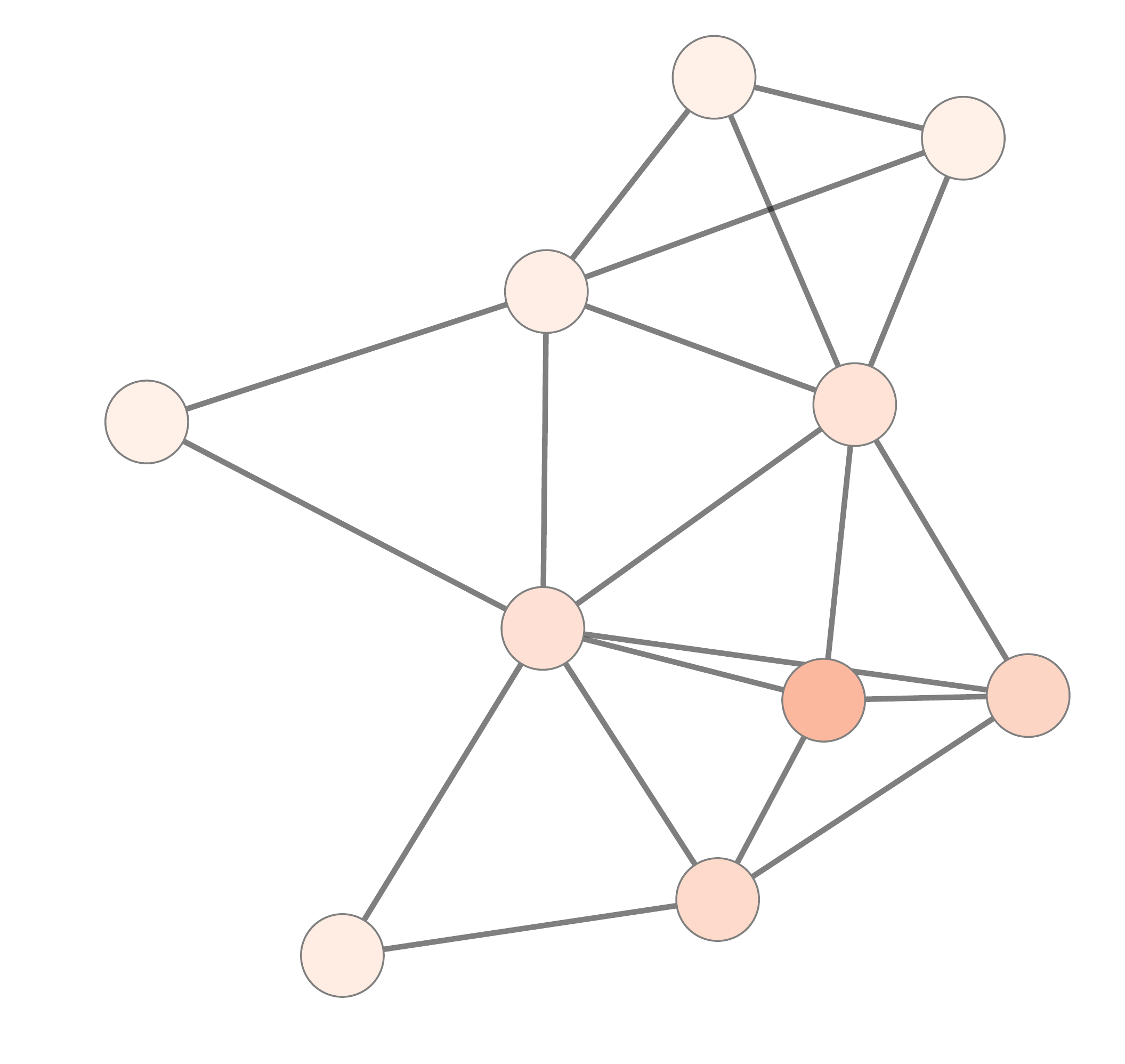}
        \caption*{$t=3$}
    \end{subfigure}
    \begin{subfigure}{0.23\linewidth}  
        \centering 
        \includegraphics[width=0.8\linewidth]{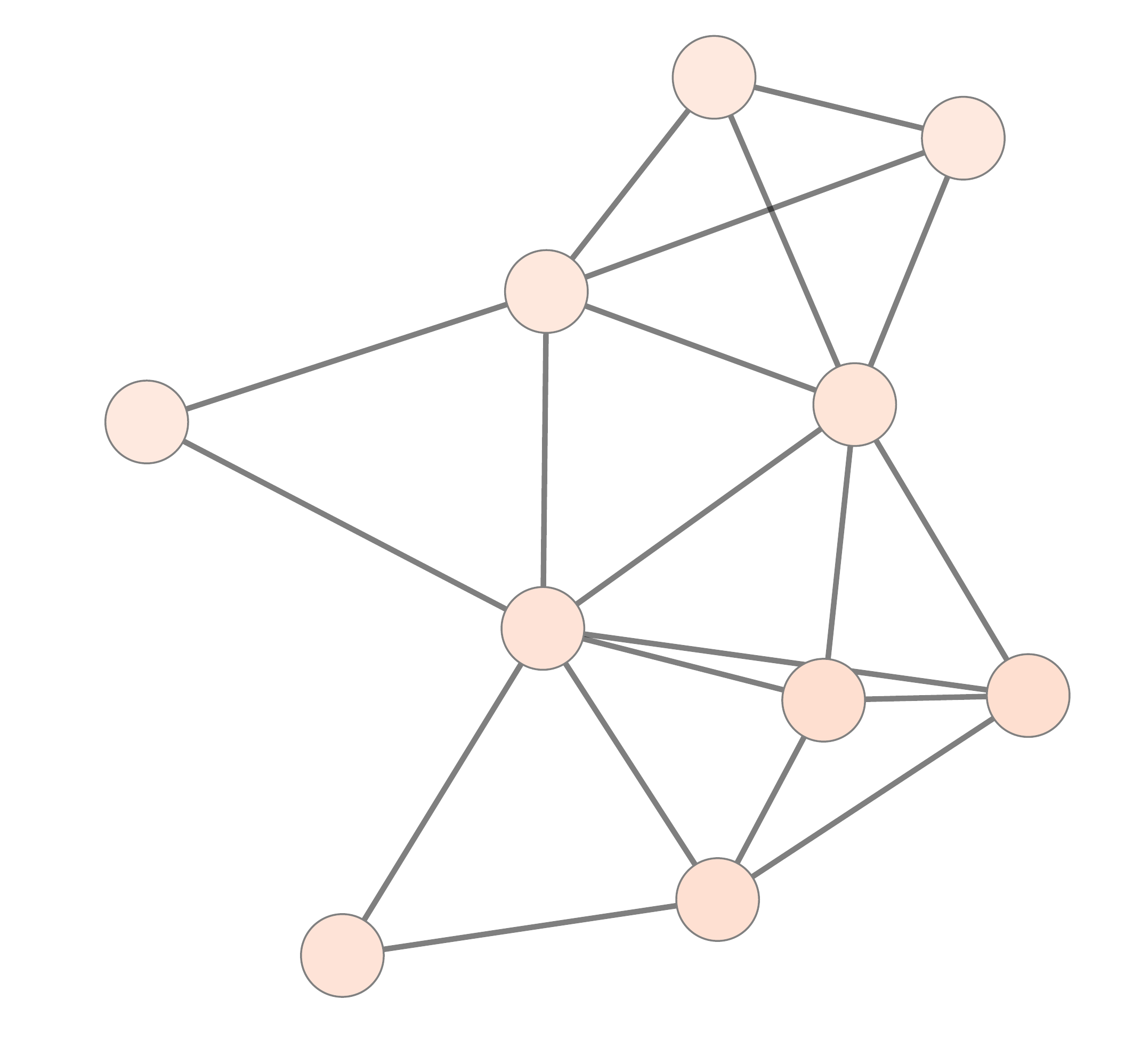}
        \caption*{$t=4$}
    \end{subfigure}
    \vskip\baselineskip
    \begin{subfigure}{0.23\linewidth}
        \centering
        \includegraphics[width=0.8\linewidth]{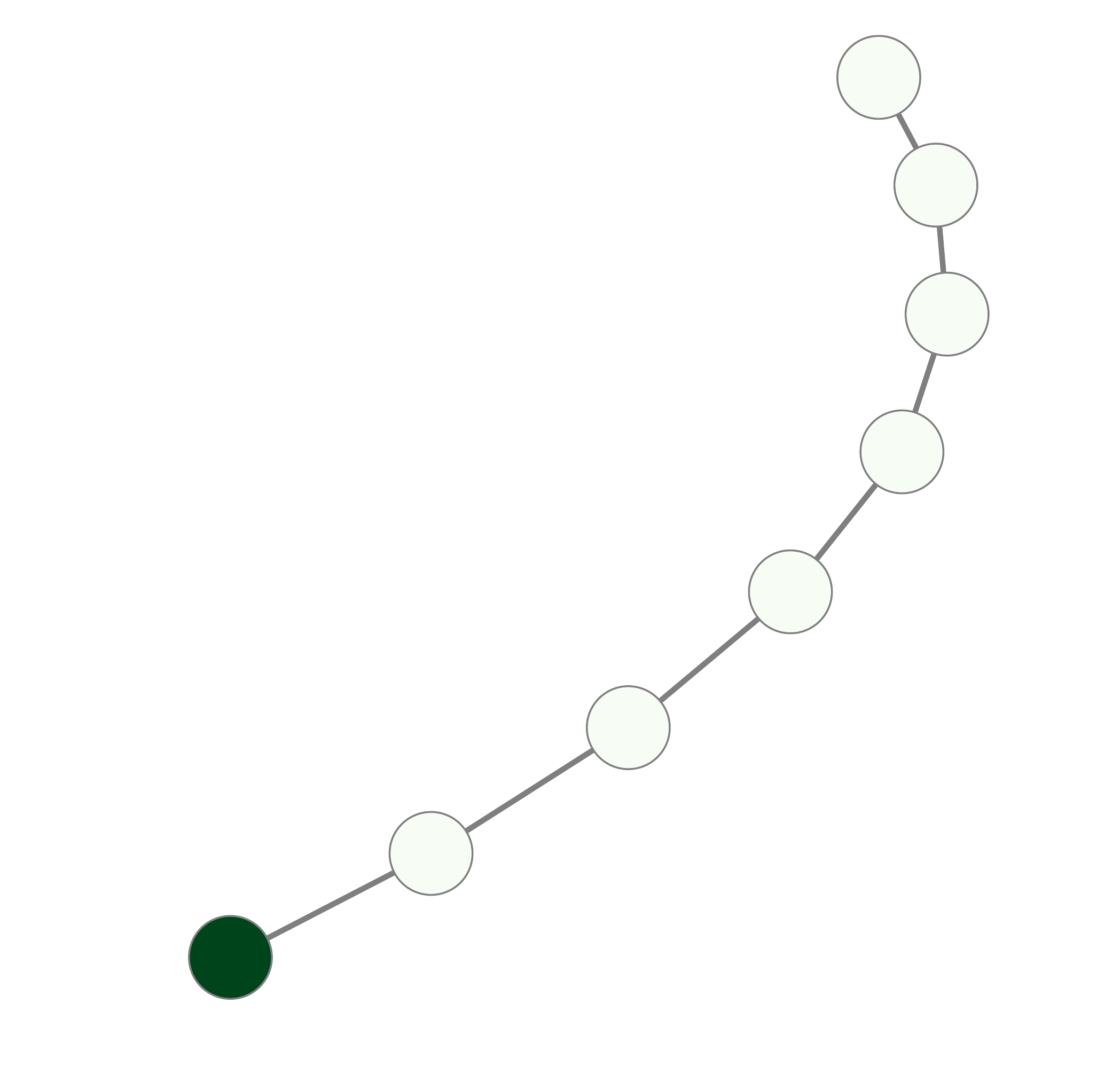}
        \caption*{$t=1$}
    \end{subfigure}
    \begin{subfigure}{0.23\linewidth}  
        \centering 
        \includegraphics[width=0.8\linewidth]{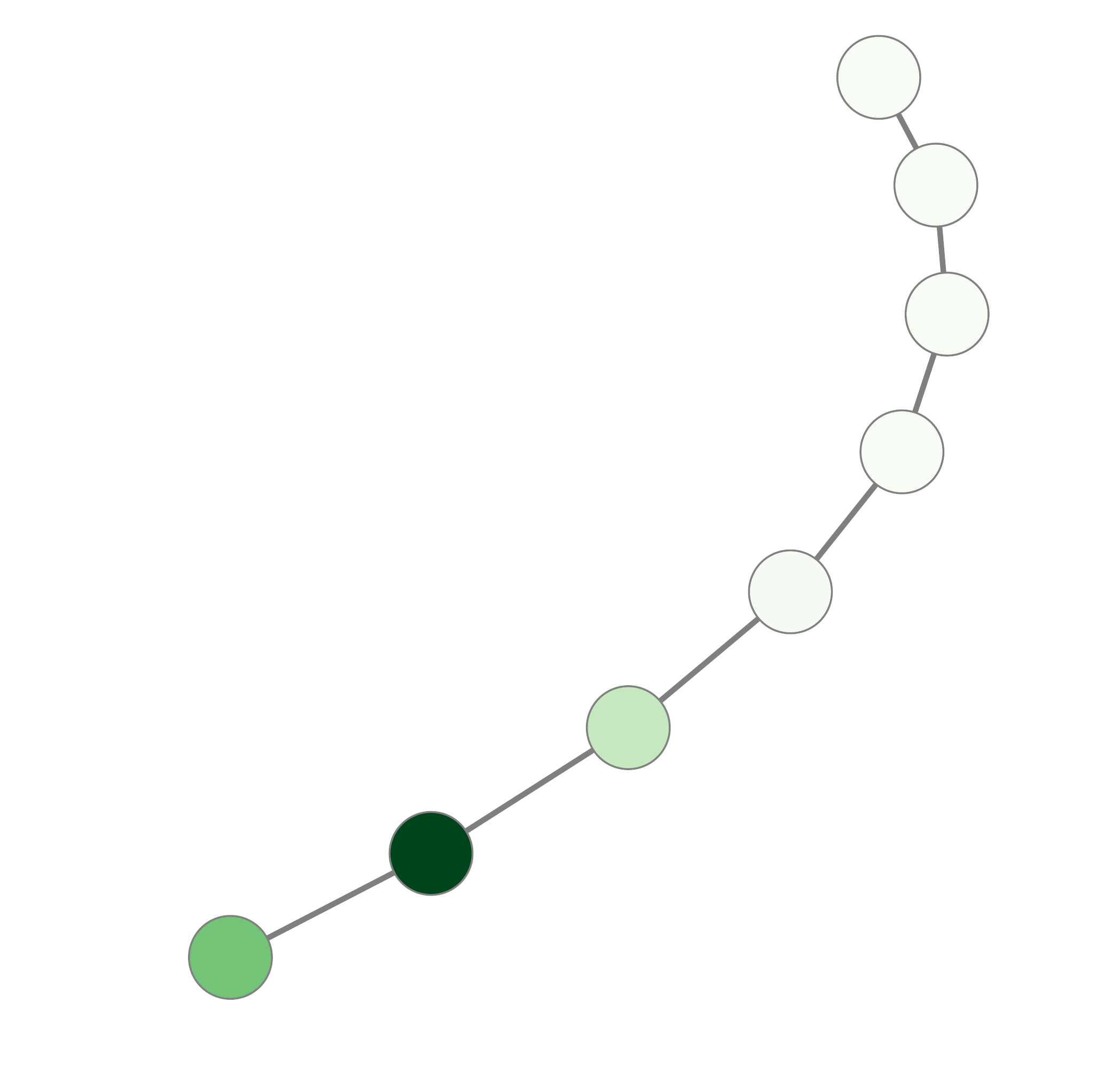}
        \caption*{$t=2$}
    \end{subfigure}
    \begin{subfigure}{0.23\linewidth}  
        \centering 
        \includegraphics[width=0.8\linewidth]{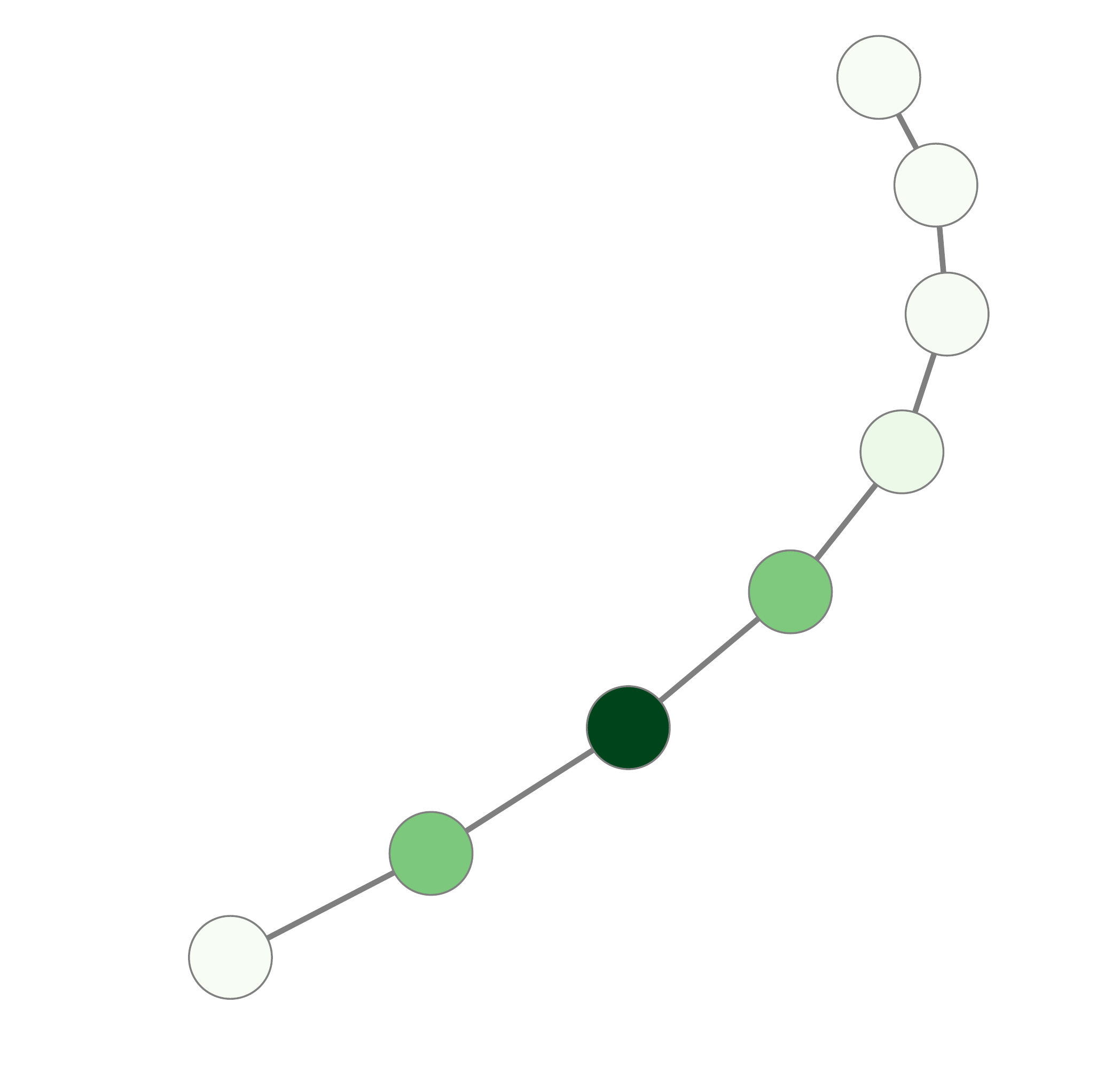}
        \caption*{$t=3$}
    \end{subfigure}
    \begin{subfigure}{0.23\linewidth}  
        \centering 
        \includegraphics[width=0.8\linewidth]{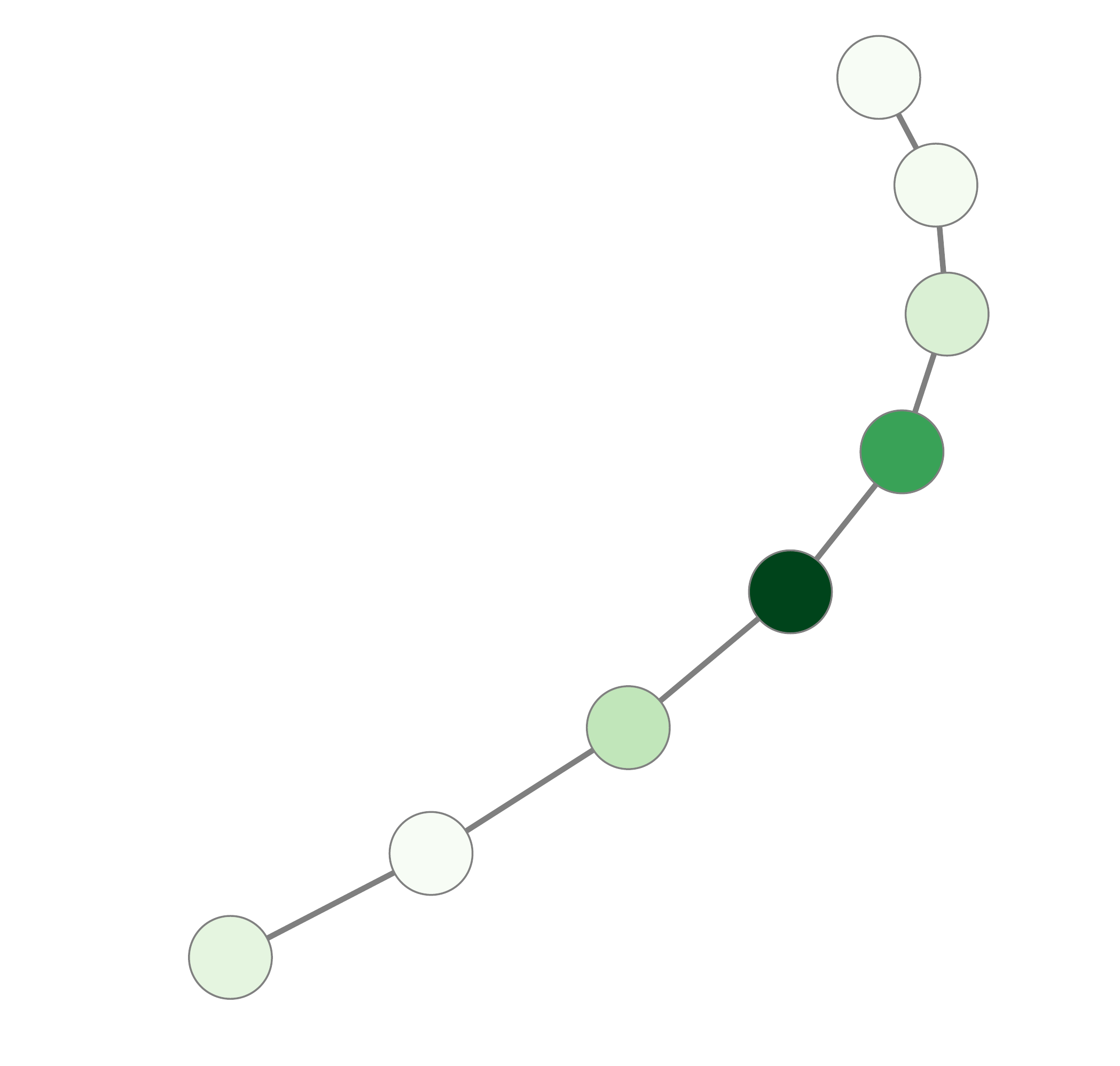}
        \caption*{$t=4$}
    \end{subfigure}
    \caption{Heat and wave dynamics on a graph.} 
    \label{fig:dynamic}
    \vskip -0.2in
\end{figure}

To visually evaluate {\ours}, we generated two sequences of snapshots based on the trained {\ours}.
Figure~\ref{fig:dynamic} illustrates how the heat (Top) on a vertex is dissipated and the pulse (Bottom) is propagated along the graph.
These sequences show that the desired dynamics can be extracted by the physics knowledge without optimizing supervised loss, and thus, it shows that physics knowledge can be beneficial.

\subsection{Climate Data}\label{sec:data}
We found that the simulated climate observations over 16 days around the Southern California region~\cite{zhang2018systematic} by using the Weather Research and Forecasting (WRF) model~\cite{skamarock2008a}.
In this dataset, the region (Latitude: 32.22 to 35.14, Longitude: -119.59 to -116.29) is divided into 18,189 grid patches and the observations are recorded hourly. We provide the details in Appendix.

\begin{minipage}{\linewidth}
    \centering
    \begin{minipage}{0.5\linewidth}
        \begin{figure}[H]
            \centering
            \includegraphics[width=0.9\linewidth]{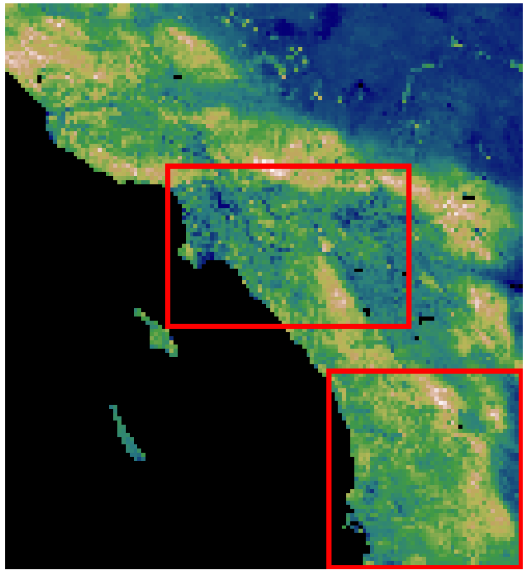}
            \caption{Southern CA region}
            \label{fig:sc-region}
        \end{figure}
    \end{minipage}
    \hspace{0.05\linewidth}
    \begin{minipage}{0.35\linewidth}
        \begin{figure}[H]
            \begin{subfigure}{0.9\linewidth}
                \centering
                \includegraphics[width=0.7\linewidth]{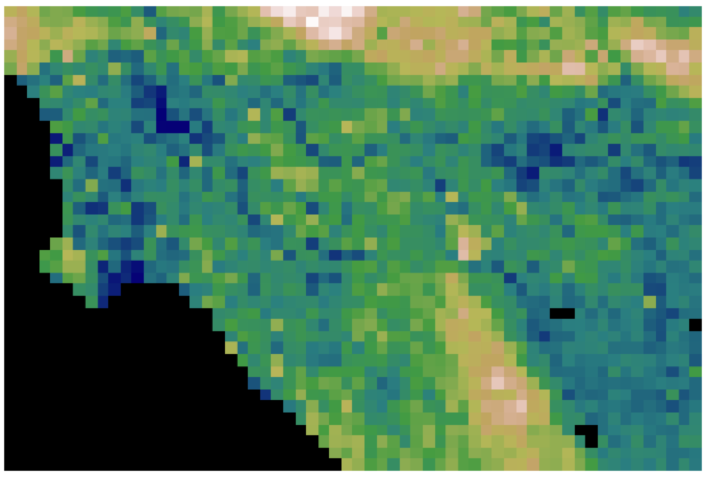}
                \caption*{LA area}
            \end{subfigure}
            \begin{subfigure}{0.9\linewidth}
                \centering
                \includegraphics[width=0.7\linewidth]{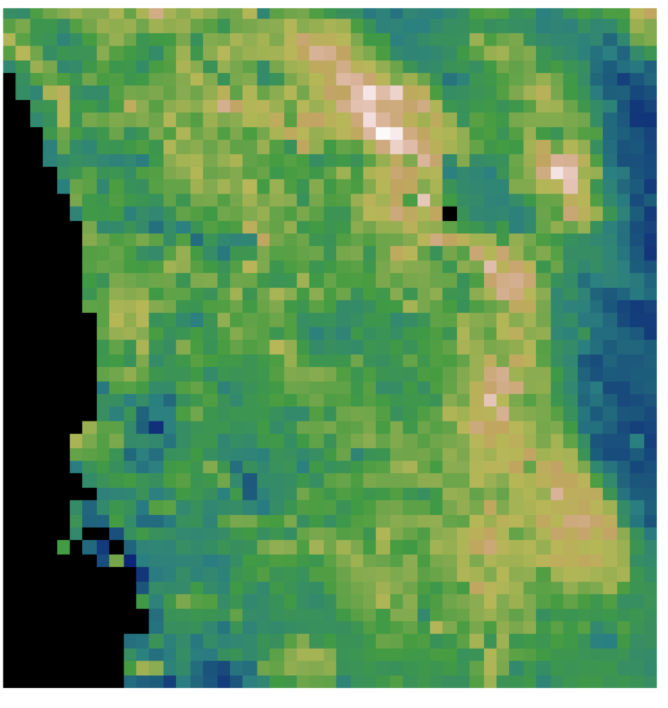}
                \caption*{SD area}
            \end{subfigure}
        \end{figure}
    \end{minipage}
\end{minipage}

Instead of using all patches at once, we sampled two subsets of the patches, Los Angeles and San Diego areas (Figure~\ref{fig:sc-region}), for training {\ours}.
To build a graph, we considered each patch as a vertex (similar to \citet{santoro2017simple}) and connect a pair of adjacent pixels to define an edge.

The vertex attributes consist of 10 climate observations, \emph{Air temperature, Albedo, Precipitation, Soil moisture, Relative humidity, Specific humidity, Surface pressure, Planetary boundary layer height, and Wind vector (2 directions)}.
Although the edge attributes are not given, we could specify the type of each edge by using the type of connected patches. 
There are 13 different land-usage types and each type summarizes how the corresponding land is used. For example, some patches are classified as commercial/industrial land (e.g., Downtown LA) but some other patches are grassland. 
Based on the type of connected patches, we assigned different embedding vectors to edges. 

\subsection{{\ours} architecture}\label{sec:arch}
As explained in Section~\ref{sec:pign}, {\ours} consists of three modules, the graph encoder, the GN block, and the graph decoder (Figure~\ref{fig:architecture}).
The encoder contains two feed forward networks, $\phi^e$ and $\phi^v$, applied to node and edge features, respectively.
By passing the encoder, the features are mapped to the latent space ($\C{H}$) where we will constrain physics equations to the hidden representations.

In the GN block, the node/edge/graph features are updated by the GN algorithm in~\citet{pmlr-v80-sanchez-gonzalez18a}. 
Here we assume that there is no curl-related constrain for modeling the climate observations.
The latent graph states, $\C{H}$ and $\C{H}'$, indicate the hidden states of the current and next observations. 
For the physics constraint, we informed the diffusion equation in Table~\ref{tab:dynamic}, which describes the behavior of the continuous physical quantities resulting from the random movement.
As the most of the climate observations are varying continuously, the diffusion equation is one of the equations that should be considered for modeling.
Note that the physics law is not directly applied to the input observations, but rather to the latent representations.
It is desired setting because it is hard to specify which observations are following the law explicitly and how much the diffusivities are. 
For example, wind vectors and surface pressure are highly probable to follow the diffusive property but they should have different behaviors.
Thus, instead of individually applying the equation to each observation, it is more efficient to introduce the constraint on the latent representations.
The state-updating process is repeated at least as many as the order of temporal derivatives in Equation~\ref{eq:dynamic} to provide the finite difference equation.
For multistep predictions, the recurrent module can be repeated more and the physics equation needs to be applied multiple times as well.

Finally, the decoder takes $\C{H}'$ as input to return the next predictions.
The following objective function is the generic total loss function of {\ours} with the diffusion equation.

\begin{align}
    \C{L} = \sum_{i=1}^T\Vert \hat{\bm{y}}_i - \bm{y}_i \Vert^2 + \lambda\sum_{i=1}^{T-M+1}\Vert \Tilde{\bm{v}}_i - \Tilde{\bm{v}}_{i-1} -\alpha\nabla^2\Tilde{\bm{v}}_{i-1} \Vert^2
    \label{eq:generic-loss}
\end{align}
where $\bm{y}$ is a vector of the target observations and $\alpha$ adjusts the diffusivity of the latent physics quantities.
Note that $\Tilde{\bm{v}}_0$ is the latent node representation of the input $\bm{v}$ and $\Tilde{\bm{v}}_{i:i>0}$ are the updated latent representations in the GN block.

\begin{table*}[t]
\centering
\caption{Prediction error (MSEs)}\label{tab:mse}
\vskip 0.1in
\begin{small}
  \begin{tabular}{ccccc}
  \cline{1-5}
    Model & \begin{tabular}[c]{@{\vspace{-0.1cm}}c@{}}LA area\\ (one-step)\end{tabular} & \begin{tabular}[c]{@{\vspace{-0.1cm}}c@{}}SD area\\ (one-step)\end{tabular}   & \begin{tabular}[c]{@{\vspace{-0.1cm}}c@{}}LA area\\ (multistep)\end{tabular} & \begin{tabular}[c]{@{\vspace{-0.1cm}}c@{}}SD area\\ (multistep)\end{tabular}   \\ \cline{1-5}
    MLP & 0.7930 $\pm$0.2327 & 1.0645 $\pm$0.2634 & - & - \\
    LSTM & 0.7378 $\pm$0.0514 & 0.9213 $\pm$0.1049 & 1.4943 $\pm$0.0970 & 1.0873 $\pm$0.0664 \\
    GN-only & 0.6035 $\pm$0.0832 & 0.7007 $\pm$0.0848 & 1.3415 $\pm$0.1195 & 1.0422 $\pm$0.0673 \\
    GN-skip & 0.56546 $\pm$0.1015 & 0.6543 $\pm$0.1195 & 1.0257 $\pm$0.1912 & 0.9872 $\pm$0.2425 \\
    {\ours} & \textbf{0.4435 $\pm$0.0378}& \textbf{0.5149 $\pm$0.0831}& \textbf{0.8677 $\pm$0.1033}& \textbf{0.6714 $\pm$0.1106} \\ \cline{1-5}
  \end{tabular}
  \end{small}
\end{table*}

\subsection{Analysis and Results}\label{sec:result}


In our experiments, we used the air temperature as a target observation and other 9 observations were used as input.
We evaluated our model by performing the one-step and multistep ($T$) prediction tasks on the two different area with a mean square error metric. 
For both tasks, we commonly fed input observations at $t$ but predicted temperature at $t+1$ or a sequence of temperature from $t+1$ to $t+T$, respectively.
First 65\% of a given sequence was used as a training set and remaining series was split into validation (10\%) and test sets (25\%).

We explored several baselines, MLP which is only able to do one-step prediction, LSTM, and GN-only that ignores the physics constraint in {\ours}.
Moreover, we compared GN-skip which connects between $\C{H}$ and $\C{H}'$ with the skip-connection~\cite{he2016deep} without the physics constraint.
All the models were trained for 10 times with randomly initialized parameters to report the MSEs with standard deviations.

Table~\ref{tab:mse} shows the prediction error of the baselines and {\ours} on different areas.
As expected, MLP and LSTM show the similar performance at the one-step prediction because the recurrent module in LSTM is practically a feed forward module in the setting.
The results from the models leveraging the given graph structure outperform MLP and LSTM. It means that knowing neighboring information is significantly helpful to infer its own state.
Among the graph-based models, {\ours} provides the least MSEs. 
It proves that it is valid reasoning to incorporate the partially given physics rule, such as the continuously varying property, with the latent representation learning.


To evaluate the effectiveness of the state-wise regularization more carefully, we conducted the multistep prediction task (10 forecast horizon).
Commonly, the models having a recurrent module are able to predict a few more steps reasonably. 
However, there are a couple of things we should pay attention.
First, the results imply that utilizing the neighboring information is important because GN-only model shows similar or better MSEs compared to LSTM for the multistep tasks, even though it has a simple recurrent module that is not as good as that of LSTM.
Second, we found that the diffusion equation in {\ours} gives the stable state transition and the property provides slowly varying latent states which are desired particularly for the climate forecasting.
Note that the skip-connection in GN-skip is also able to restrict the rapid changes of $\C{H}$.
However, it is necessary to more carefully optimize the parameters in GN-skip to learn the residual term in $\C{H}'=\C{H}+\text{GN}(\C{H})$ properly.

\subsection{Effectiveness of Physics Constraint}
\begin{figure}[t]
    \centering
    \begin{subfigure}{0.47\linewidth}  
        \centering 
        \includegraphics[width=1.0\linewidth]{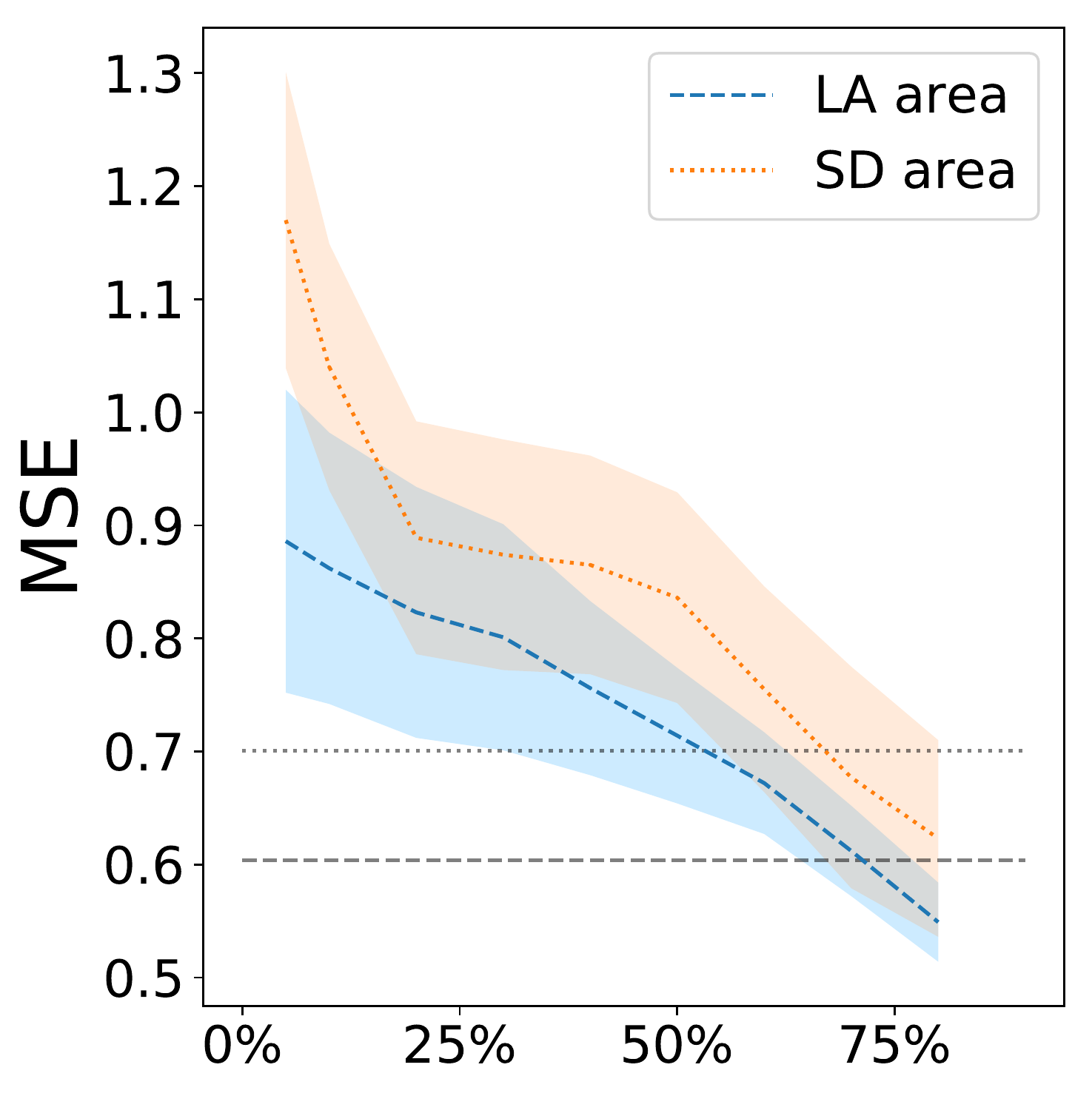}
        \caption{MSEs on sampled data}\label{fig:eff-phy}
    \end{subfigure}
    \begin{subfigure}{0.47\linewidth}
        \centering
        \includegraphics[width=1.0\linewidth]{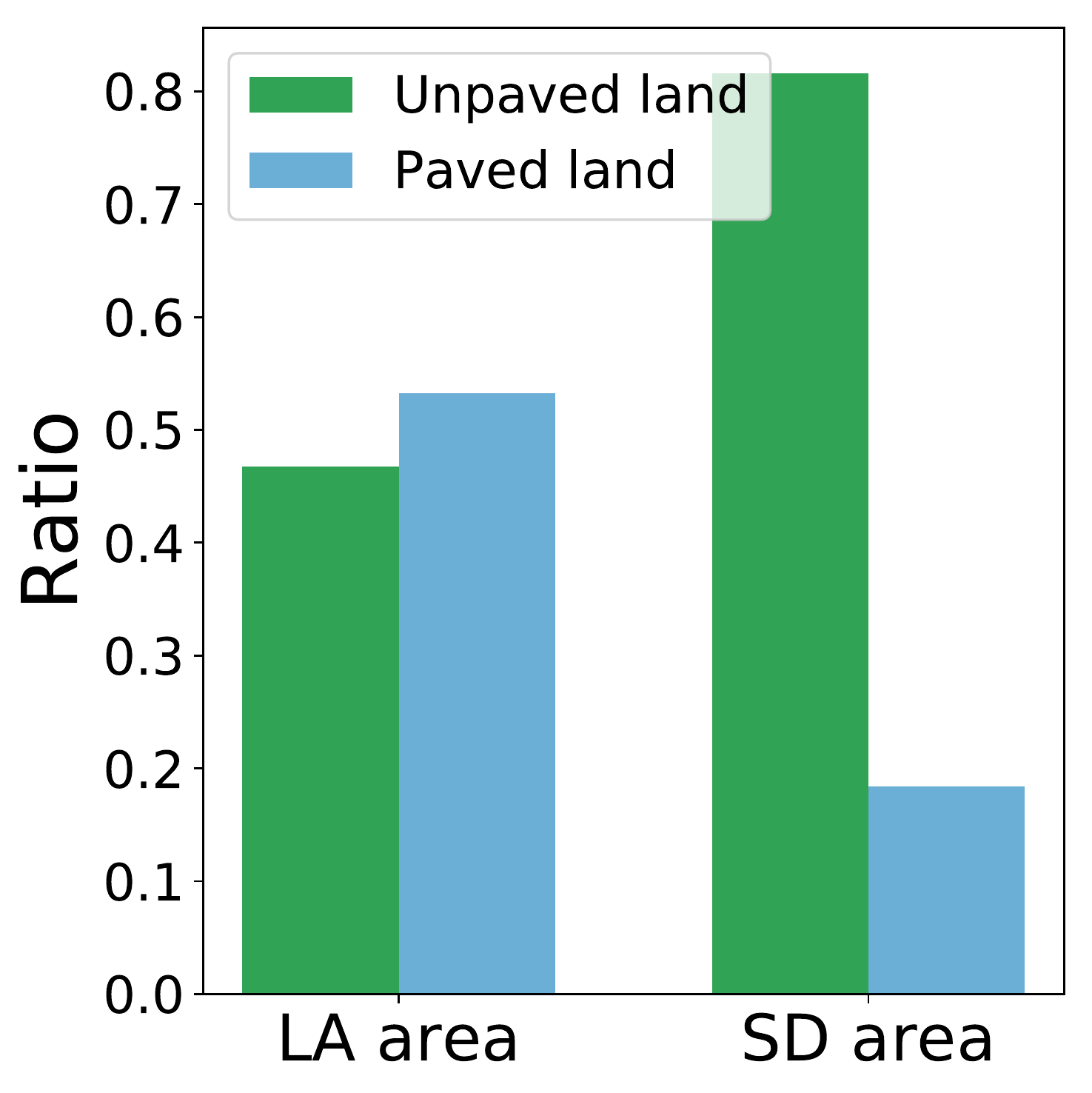}
        \caption{Ratio of land types}\label{fig:ratio}
    \end{subfigure}
    \caption{In (a), MSEs of {\ours} are almost as good as GN-only (gray lines) even though partial training data are used for the supervised loss. (b) illustrates the land usage distribution of each area. Unlike LA area, SD area has a less portion of paved land such as commercial or residential land. Thus, climate changing in an urban area might not be properly modeled with that area.}
    \vskip -0.2in
\end{figure}

We next explore how much the physics constraint is helpful by testing whether {\ours} can be trained reasonably when the number of data for the supervised objective is limited for the one-step prediction task.
We randomly sampled training data which were used to optimize the total loss function (Equation~\ref{eq:generic-loss}) and the left unsampled data were only used to minimize the physics constraint: 
\begin{align*}
    & \C{L} = \C{L}^i_{sup} + \lambda\C{L}^i_{phy}, \quad \text{$i$ is a sampled step}\\
    &\C{L} = \lambda\C{L}^i_{phy}, \qquad\qquad \text{otherwise}
\end{align*}
We find that the diffusion equation can benefit to optimize {\ours} even if the target observations are partially available (Figure~\ref{fig:eff-phy}).
Although the overall performances of {\ours} are degraded when less number of sampled data are used, the error are not far deviated from those of GN-only.
Even the GN-only model is outperformed by {\ours} when only 70\% training data are used with the state-wise constraint


\subsection{Inductive Prediction}
In {\ours}, the node- and edge-centric functions are permutationally invariant.
In other words, these functions are not particularly dependent on an order of nodes or edges, instead, they more generalize relations among the components in a graph, thus it is easier to extend to an inductive framework, inferring relations between unseen nodes and edges.
As presented in~\citet{hamilton2017inductive}, we explore if {\ours} can benefit successfully inductive representations.
We train {\ours} on the LA area only and tested it on the SD area and vice versa.
For some edge attributes in a region not found in another region, we assign a constant embedding vector to the edge attribute.
Table~\ref{tab:inductive} presents the results of the inductive prediction task.

\begin{table}[t]
    \centering
    \small
    \caption{Inductive prediction error (MSEs)}\label{tab:inductive}
    \vskip 0.1in
    \begin{tabular}{ccccc}
    \cline{1-5}
        Model & \begin{tabular}[c]{@{\vspace{-0.1cm}}c@{}}LA$\rightarrow$SD\\ (one-step)\end{tabular} &  \begin{tabular}[c]{@{\vspace{-0.1cm}}c@{}}LA$\rightarrow$SD\\ (multistep)\end{tabular} & 
        \begin{tabular}[c]{@{\vspace{-0.1cm}}c@{}}SD$\rightarrow$LA \\ (one-step)\end{tabular} & \begin{tabular}[c]{@{\vspace{-0.1cm}}c@{}}SD$\rightarrow$LA \\ (multistep)\end{tabular} \\ \cline{1-5} 
        GN-only & 0.6963 & 1.2445  & 0.8597 & 1.5170\\
        {\ours} & 0.5729 & 0.7998  & 0.7222 & 1.1055\\ \cline{1-5}
    \end{tabular}
    \vskip -0.2in
\end{table}

According to the results, we can find that GN-only and {\ours} can be generalized to learn the common patterns in the separate sub-regions.
Interestingly, the one-step prediction error from LA to SD are close to the supervised error on SD area (Table~\ref{tab:mse}).
The finding implies that the dependencies between a current and next climate state in both regions are similar and thus, the inductive learning works. 
However, it is not very effective to predict temperature on the LA area by using the trained model on the SD area. 
It is because the ratios of the land types on both regions are remarkably different.
As Figure~\ref{fig:ratio} illustrates, the LA area uniformly consists of unpaved patches and paved patches which cause pretty different climate observations and behaviors, respectively.
On the other hand, the SD area has unpaved patches mostly.
Thus, the model on the SD area is more likely biased to the unpaved characteristics and it causes the higher inductive error when it is applied to infer the relations between paved lands in the LA area.

\subsection{Robustness of Longer Predictions}
\begin{figure}[t]
    \centering
    \begin{subfigure}{0.47\linewidth}
        \centering
        \includegraphics[width=1.0\linewidth]{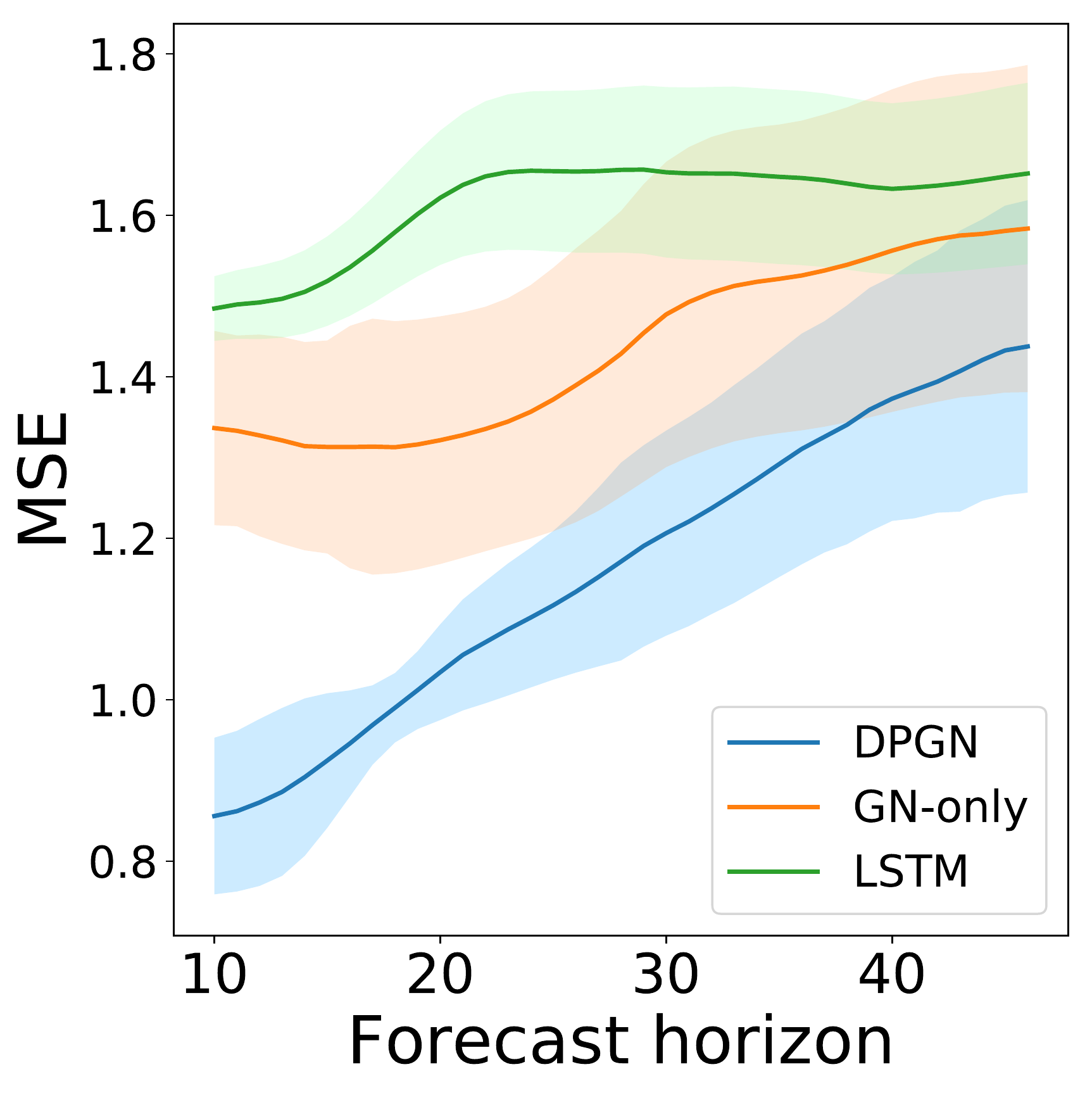}
        \caption{LA area}\label{fig:hor-la}
    \end{subfigure}
    \begin{subfigure}{0.47\linewidth}  
        \centering 
        \includegraphics[width=1.0\linewidth]{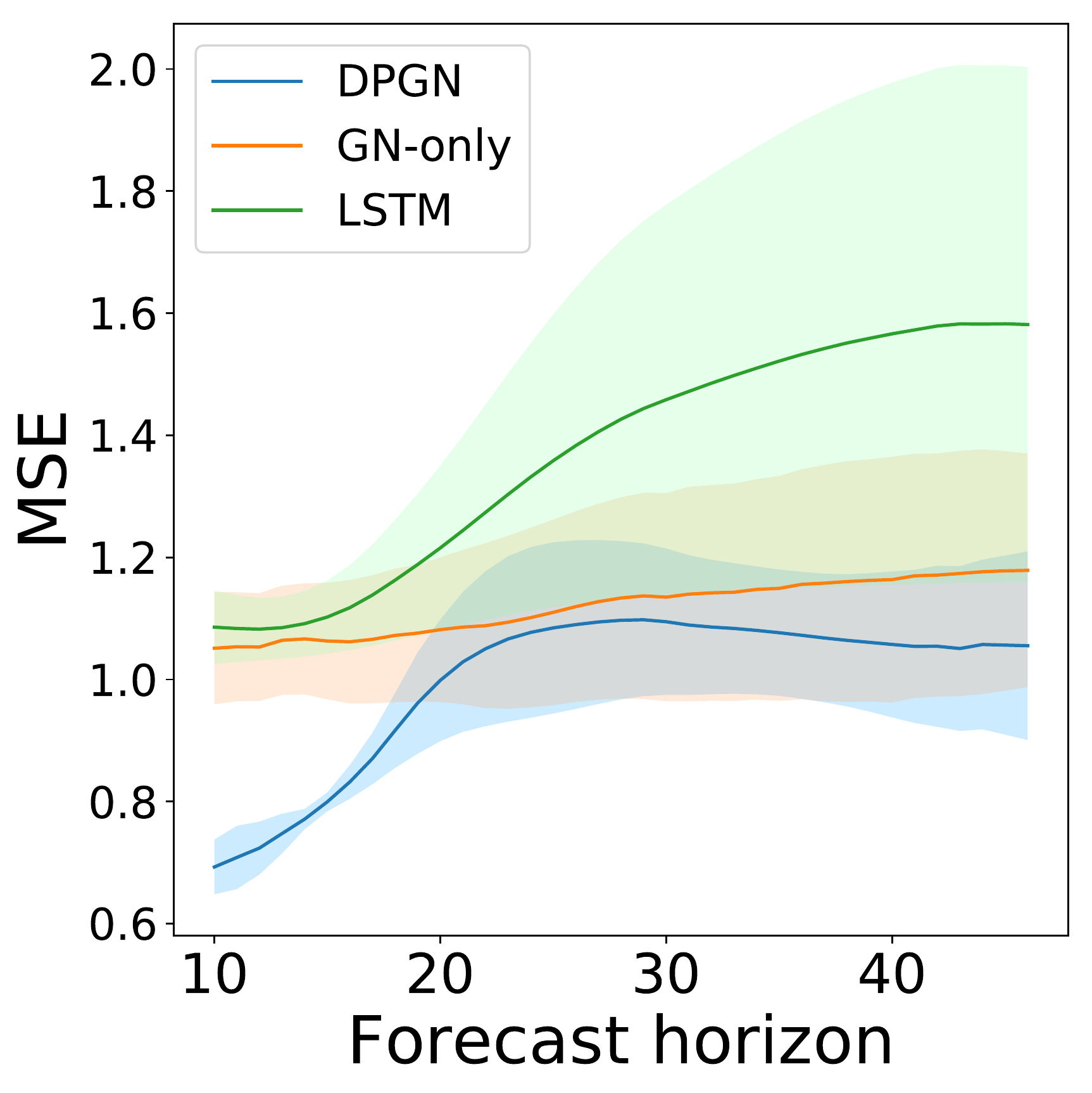}
        \caption{SD area}\label{fig:hor-sd}
    \end{subfigure}
    \caption{MSE versus the forecast horizons for the three models with standard deviations.}\label{fig:hor}
    \vskip -0.2in
\end{figure}

To evaluate if {\ours} can be a potential tool for practically simulating climate observations, we explore how much longer predictions are robust.
To utilize the recurrent module, we train each model with 10 forecasting horizons and tested over 10 to 47 horizons.
Figure~\ref{fig:hor-la} and~\ref{fig:hor-sd} illustrate how the error are varied over the different forecasting horizons in the LA area and the SD area, respectively.
Three models are robust when they forecast only few more steps ahead, however, compared to other two, {\ours} provides the most stable and accurate predictions.
As the forecast horizon increases, these models start to present larger and more varying error.
Particularly, GN-only is getting more and more unstable (wider standard deviations) as it predicts further horizons compared to other two models in the LA area (Figure~\ref{fig:hor-la}). 
On the other hand, we found the relatively stable error from GN-only and {\ours} over the SD area (Figure~\ref{fig:hor-sd}). 
The results come from the fact that the temperature observations over time in the SD area are not fluctuating as much as those in the LA area.
In other words, although it is a bit harder to predict next temperature from current observations in the SD area, it is relatively stable to predict temperature at more forecast horizons due to the less fluctuation in the SD area.

\section{Conclusion}
In this work, we introduce a new architecture based on graph networks to incorporate prior knowledge given as a form of PDEs over time and space.
We first provide how the graph networks framework generalize the differential operators in a graph.
Then, we presente a regularization which is a function of consecutive latent states and spatial differences of the states to inject a given physics equation.

While existing works more focus on how to extract useful representations from data generated by explicit physics rules, we propose a method to utilize generally required knowledge in data.
We examine if the spatiotemporal constraint is valid across a range of experiments on the climate observations.

Despite of the experiments results, there remains several topics to be studied as future works.
First, it is an interesting question whether {\ours} is still valid when physics knowledge is given as more complicated multiple PDEs.
Second, it is also necessary to explore if {\ours} is applicable to other challenging learning scenarios, noisy data or fewer labeled examples.
Moreover, it is worth to explore how to make {\ours} more stable and robust in long range predictions.


\newpage
\small
\bibliography{references}
\bibliographystyle{icml2019}

\newpage
\appendix
\section{Appendix}

\subsection{Climate Data}

In this work, we used the simulated climate dataset published in~\cite{zhang2018systematic}. 
In the dataset, a number of climate observations are recorded based on the conventional WRF model.
The observations cover the period from June/28/2012 21:00 to July/14/2012 20:00 over the Southern California region (Latitude: 32.226044 to 35.140636 and Longitude: -119.593155 to -116.29416).

\begin{figure}[h]
    \centering
    \includegraphics[width=0.8\linewidth]{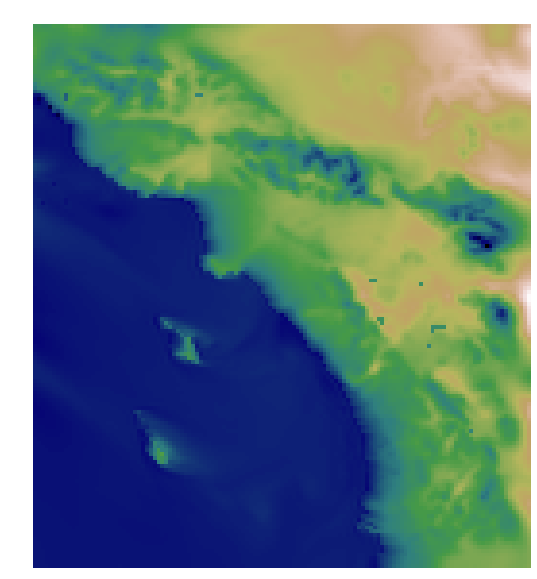}
    \caption{Data domain}
    \label{fig:sc-map}
\end{figure}

\begin{figure}[t]
    \centering
    \begin{subfigure}{0.23\linewidth}
        \centering
        \includegraphics[width=0.8\linewidth]{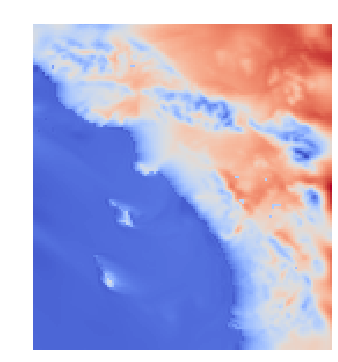}
        \caption*{$t=1$}
    \end{subfigure}
    \begin{subfigure}{0.23\linewidth}  
        \centering 
        \includegraphics[width=0.8\linewidth]{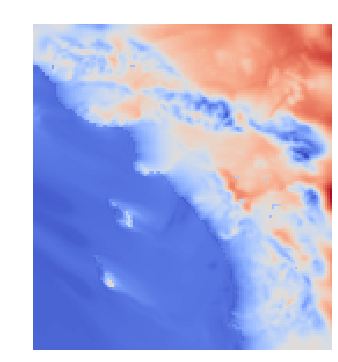}
        \caption*{$t=2$}
    \end{subfigure}
    \begin{subfigure}{0.23\linewidth}  
        \centering 
        \includegraphics[width=0.8\linewidth]{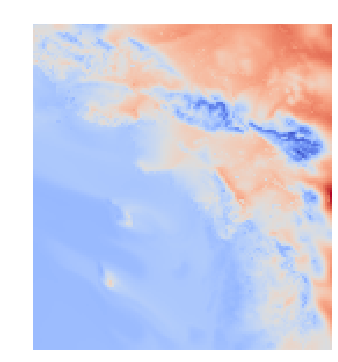}
        \caption*{$t=3$}
    \end{subfigure}
    \begin{subfigure}{0.23\linewidth}  
        \centering 
        \includegraphics[width=0.8\linewidth]{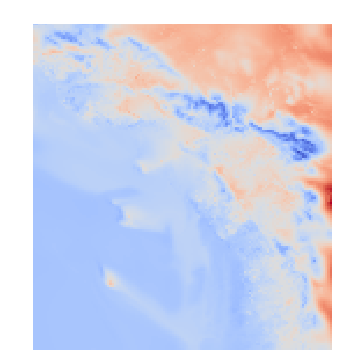}
        \caption*{$t=4$}
    \end{subfigure}
    \vskip\baselineskip
    \begin{subfigure}{0.23\linewidth}
        \centering
        \includegraphics[width=0.8\linewidth]{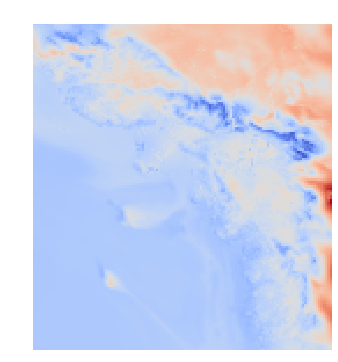}
        \caption*{$t=5$}
    \end{subfigure}
    \begin{subfigure}{0.23\linewidth}  
        \centering 
        \includegraphics[width=0.8\linewidth]{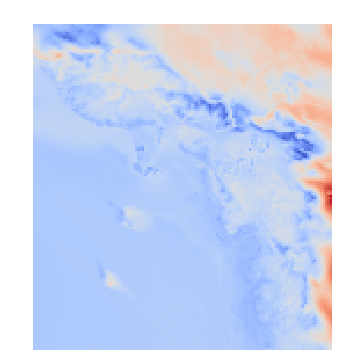}
        \caption*{$t=6$}
    \end{subfigure}
    \begin{subfigure}{0.23\linewidth}  
        \centering 
        \includegraphics[width=0.8\linewidth]{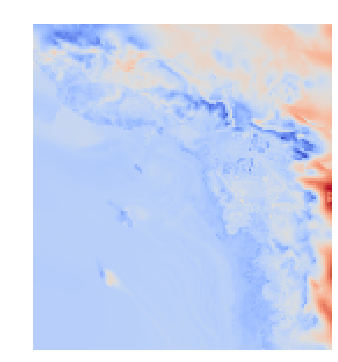}
        \caption*{$t=7$}
    \end{subfigure}
    \begin{subfigure}{0.23\linewidth}  
        \centering 
        \includegraphics[width=0.8\linewidth]{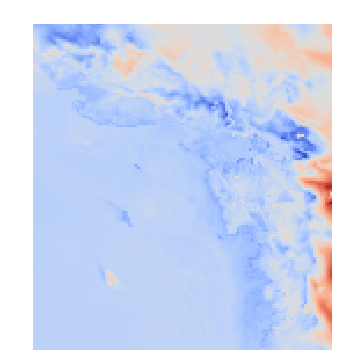}
        \caption*{$t=8$}
    \end{subfigure}
    \vskip\baselineskip
    \begin{subfigure}{0.23\linewidth}
        \centering
        \includegraphics[width=0.8\linewidth]{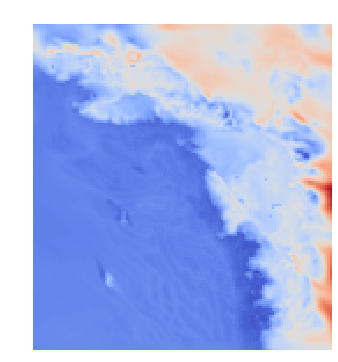}
        \caption*{$t=9$}
    \end{subfigure}
    \begin{subfigure}{0.23\linewidth}  
        \centering 
        \includegraphics[width=0.8\linewidth]{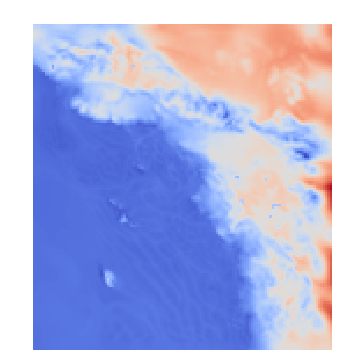}
        \caption*{$t=10$}
    \end{subfigure}
    \begin{subfigure}{0.23\linewidth}  
        \centering 
        \includegraphics[width=0.8\linewidth]{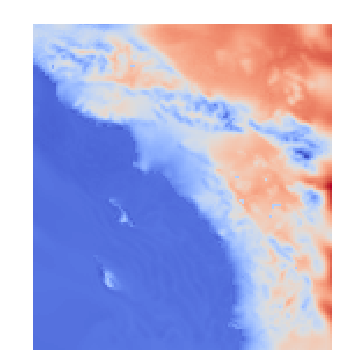}
        \caption*{$t=11$}
    \end{subfigure}
    \begin{subfigure}{0.23\linewidth}  
        \centering 
        \includegraphics[width=0.8\linewidth]{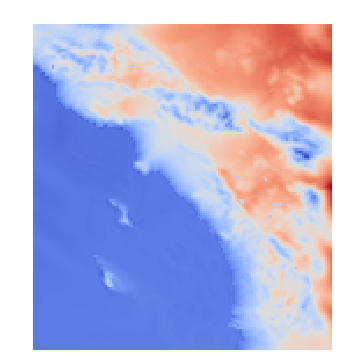}
        \caption*{$t=12$}
    \end{subfigure}
    \caption{Temperature observations over 24 hours} 
    \label{fig:temps}
    \vskip -0.2in
\end{figure}

Here we provide the details for the climate observations and basic information.

\begin{table}[h]
    \centering
    \caption{Functions in {\ours}}
    \label{tab:archi}
    \vskip 0.1in
    \begin{tabular}{c|c}
    \cline{1-2}
         &  hyper-parameter \\ \cline{1-2}
        Node encoder & MLP [9,64] with ReLU\\
        Edge encoder & Embedding matrix \\
        Edge update func, $\phi^e$ & \begin{tabular}[c]{@{}l@{}}$\bm{e}'_{ij}=\phi^e(\bm{e}_{ij},\bm{v}_i,\bm{v}_j,\bm{u})$\\ MLP [256,64] with ReLU \end{tabular} \\ 
        Node update func, $\phi^v$ & \begin{tabular}[c]{@{}l@{}}$\bm{v}'_{i}=\phi^v(\bm{v}_i,\bar{\bm{e}}'_{i\cdot},\bar{\bm{e}}'_{\cdot i},\bm{u})$\\ MLP [256,64] with ReLU\end{tabular} \\ 
        Global update func, $\phi^u$ & \begin{tabular}[c]{@{}l@{}}$\bm{u}'=\phi^u(\bm{u},\bar{\bm{e}}',\bar{\bm{v}}')$\\ MLP [192,64] with ReLU \end{tabular} \\
        Edge aggregator for $\bm{v}$ & $\rho^e(\bm{e}_i)=\text{Add}(\bm{e}_i)$ \\
        Edge aggregator for $\bm{u}$ & $\rho^e(\bm{e}_i)=\text{Avg}(\bm{e}_i)$ \\
        Node aggregator for $\bm{u}$ & $\rho^v(\bm{v}_i)=\text{Avg}(\bm{v}_i)$ \\
        Node decoder & MLP [64,1] \\
        \cline{1-2}
    \end{tabular}
\end{table}

\begin{table*}[h]
    \centering
    \caption{Description of climate data}
    \label{tab:description}
    \vskip 0.1in
    \begin{tabular}{c|l}
    \cline{1-2}
        Feature &  Description\\ \cline{1-2}
        Timestamp & Every 60 minute \\
        T2 (3D) & Near-surface (2 meter) air temperature (unit: $K$) (time varying)\\
        XLAT (3D) & Latitude (unit: degree north) (time invariant) \\
        XLONG (3D) & Longitude (unit: degree east)  (time invariant) \\
        ALBEDO (3D) & Albedo (unit: -) (time varying) \\
        FRC\_URB2D (3D) & Impervious fraction, urban fraction (unit: -) (time invariant) \\
        VEGFRA (3D) & Vegetation fraction (unit: -) (time invariant) \\
        LU\_INDEX (3D) & \begin{tabular}[c]{@{}l@{}}Land use classification (unit: -) (time invariant)\\ \small Paved index: (31: Low-intensity residential,  32: High-intensity residential,  33: Commercial/Industrial) \end{tabular} \\ 
        U (4D) & Wind vector, x-wind component (unit: $m*s^{-1}$) (west to east vector) (time varying)\\ 
        V (4D) & Wind vector, y-wind component (unit: $m*s^{-1}$) (south to north vector) (time varying)\\
        RAINNC (3D) & Accumulated total grid scale precipitation (unit: mm) (time varying)\\
        SMOIS (4D) & Soil moisture (unit: $m^3 m^{-3}$) (time varying)\\
        PBLH (3D) & Boundary layer height (unit: $m$) (time varying)\\
        RH2 (3D) & 2-meter relative humidity (unit: -) (time varying)\\
        Q2 (3D) & 2-meter specific humidity (units: $kg kg^{-1}$) (time varying)\\
        PSFC (3D) & surface pressure (units: Pa) (time varying)\\
        \cline{1-2}
    \end{tabular}
\end{table*}

\begin{table}[h]
    \centering
    \caption{Description of sub-regions}
    \label{tab:sub-region}
    \vskip 0.1in
    \begin{tabular}{c|cc}
    \cline{1-3}
         &  LA area & SD area\\ \cline{1-3}
        \# of total patches & 2400 & 2499\\ 
        \# of sampled patches & 280 & 272 \\
        \# unpaved patches & 130 & 217 \\
        \# paved patches & 148 & 49 \\
        \# unknown patches & 2 & 6\\
        \# edge attributes & 45 & 42 \\
        \cline{1-3}
    \end{tabular}
\end{table}

\newpage
\subsection{Model Details}
Here we provide additional details for all models we used in the work, including the exact hyper-parameter and architecture settings.
All models were trained using the Adam optimiser, with exponential decay rate parameters $\beta_1=0.9$, $\beta_2=0.999$, $\epsilon=10^{-8}$. 
All experiments were done on GeForce GTX 1080 Ti.

\begin{table}[h]
    \centering
    \caption{Hyper-parameter of {\ours}}
    \label{tab:hyper}
    \vskip 0.1in
    \begin{tabular}{c|c}
    \cline{1-2}
         &  hyper-parameter \\ \cline{1-2}
        Edge embedding dim & 64 \\ 
        Edge hidden dim & 64\\
        Node hidden dim & 64\\
        Global hidden dim & 64\\
        Learning rate & 0.001 \\
        Iterations & 30,000 \\
        $\lambda$ & 1e-5\\
        Diffusion coefficient & 0.001 \\
        \cline{1-2}
    \end{tabular}
\end{table}





\end{document}